%% file: templateArxiv.tex
\documentclass{article}

\usepackage{PRIMEarxiv}

\usepackage[utf8]{inputenc} % allow utf-8 input
\usepackage[T1]{fontenc}    % use 8-bit T1 fonts
\usepackage[hidelinks]{hyperref}
\usepackage{orcidlink}
\usepackage{url}            % simple URL typesetting
\usepackage{booktabs}       % professional-quality tables
\usepackage{amsfonts}       % blackboard math symbols
\usepackage{nicefrac}       % compact symbols for 1/2, etc.
\usepackage{microtype}      % microtypography
\usepackage{lipsum}
\usepackage{fancyhdr}       % header
\usepackage{graphicx}       % graphics

\usepackage{xcolor}
\usepackage{colortbl}
\usepackage{csquotes}
\usepackage{mathtools}
\usepackage{dsfont}
\usepackage{multirow}
\usepackage{multicol}
\usepackage{pifont} % for checkmark and x-mark
\usepackage{array} % centered text in table cells
\newcolumntype{M}[1]{>{\centering\arraybackslash}m{#1}}
\usepackage{adjustbox} % for rotated table headers
\usepackage{diagbox} % diagonally split table cell
\usepackage[nocompress]{cite} % for ordered citations [2,1,4] -> [1,2,4]

\graphicspath{{Images/}{target_progression/}{perturbations_universal/}{targets_0.1/}{perturbation_variations}} 
\usepackage{bibunits} % Allows restarting the bibliography counter for the supplementary material

\newcommand{\cf}{\emph{cf.}\ }
\newcommand{\etal}{\emph{et al.}}

\newcommand{\Rds}{\mathds{R}}
\newcommand{\targ}{f^t}
\newcommand{\fadv}{\check{f}}
\newcommand{\flow}{f}
\newcommand{\fgt}{f^g}
\newcommand{\ltwo}{$L_2$}
\newcommand{\linfty}{$L_\infty$}
\newcommand{\lzero}{$L_0$}
\newcommand{\dd}{\delta_t, \delta_{t+1}}
\newcommand{\dc}{\delta_{t,t+1}}
\newcommand{\ddu}{\overline{\dd}}
\newcommand{\dcu}{\overline{\dc}}
\newcommand{\img}{\mathcal{I}}
\newcommand{\loss}{\mathcal{L}}

\newcommand{\yes}{\text{\ding{51}}}%
\newcolumntype{R}[2]{%
    >{\adjustbox{angle=#1,lap=\width-(#2)}\bgroup}%
    l%
    <{\egroup}%
}
\newcommand{\executeiffilenewer}[3]{\ifnum\pdfstrcmp{\pdffilemoddate{#1}}%
      {\pdffilemoddate{#2}}>0%
      {\immediate\write18{#3}}\fi%
}

\makeatletter
\newcommand{\settitle}{\@maketitle}
\makeatother

\title{A Perturbation-Constrained Adversarial Attack for Evaluating the Robustness of Optical Flow
}

\author{
  Jenny Schmalfuss \orcidlink{0000-0001-8507-927X}\hspace{20mm} Philipp Scholze \orcidlink{0000-0003-0776-0144}\hspace{20mm} Andrés Bruhn \orcidlink{0000-0003-0423-7411}\\[3mm]
  Institute for Visualization and Interactive Systems\\
  University of Stuttgart \\
  \texttt{\{jenny.schmalfuss,andres.bruhn\}@vis.uni-stuttgart.de} \\
   \texttt{philipp.scholze@simtech.uni-stuttgart.de} \\
}

\begin{document}
% \maketitle
\settitle

\begin{abstract}
Recent optical flow methods are almost exclusively judged in terms of accuracy, while their robustness is often neglected.
Although adversarial attacks offer a useful tool to perform such an analysis, current attacks on optical flow methods focus on real-world attacking scenarios rather than a \emph{worst case} robustness assessment.
Hence, in this work, we propose a novel adversarial attack -- the Perturbation-Constrained Flow Attack (PCFA) --  that emphasizes destructivity over applicability as a real-world attack.
PCFA is a global attack that optimizes adversarial perturbations to shift the predicted flow towards a specified target flow, while keeping the \ltwo\ norm of the perturbation below a chosen bound.
Our experiments demonstrate PCFA's applicability in white- and black-box settings, and show it finds stronger adversarial samples than previous attacks.
Based on these strong samples, we provide the first joint ranking of optical flow methods considering both prediction quality and adversarial robustness, which reveals state-of-the-art methods to be particularly vulnerable. Code is available at \href{https://github.com/cv-stuttgart/PCFA}{https://github.com/cv-stuttgart/PCFA}.
\end{abstract}

% keywords can be removed
%\keywords{Optical Flow \and Robustness \and Global Adversarial Attack \and \ltwo~Constrained Perturbation}

\begin{center}
\end{center}

\section{Introduction}
\begin{figure}[b]
\centering
\setlength{\fboxrule}{0.1pt}%
\setlength{\fboxsep}{0pt}%
\small
\begin{tabular}{@{}r@{\ }M{36mm}@{\ \ }M{36mm}@{\ }M{36mm}@{\ }M{36mm}@{}}
            & \small Initial, $\varepsilon_2=0$ & \small $\varepsilon_2 \leq 5\cdot 10^{-4}$ & \small $\varepsilon_2 \leq 5\cdot 10^{-3}$ & \small $\varepsilon_2 \leq 5\cdot 10^{-2}$ \\
   \small Flow $\flow$ & \fcolorbox{gray}{white}{\includegraphics[width=36mm]{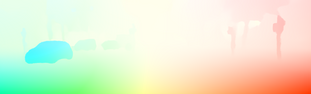}} & \fcolorbox{gray}{white}{\includegraphics[width=36mm]{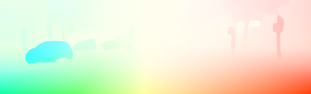}} & \fcolorbox{gray}{white}{\includegraphics[width=36mm]{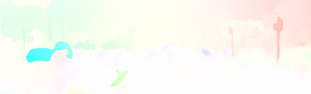}} & \fcolorbox{gray}{white}{\includegraphics[width=36mm]{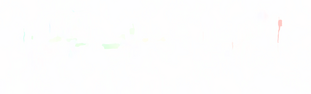}}  \\
   \small Pert. Inp. $\mathcal{I}_t$ & \includegraphics[width=36mm]{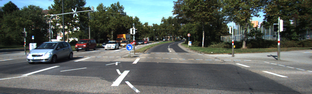} & \includegraphics[width=36mm]{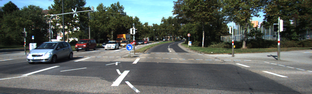} & \includegraphics[width=36mm]{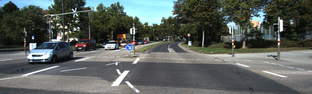} & \includegraphics[width=36mm]{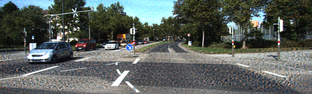}  \\
   \small Perturb. $\delta_t$ & \includegraphics[width=36mm]{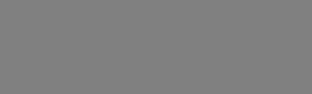} & \includegraphics[width=36mm]{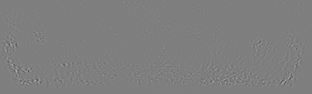} & \includegraphics[width=36mm]{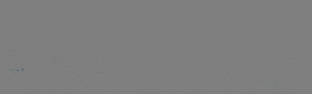} & \includegraphics[width=36mm]{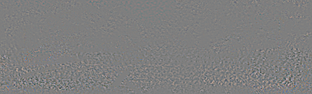}  \\
\end{tabular}
\caption{Robustness evaluation for RAFT~\cite{teed2020raft}. Here, our Perturbation-Constrained Flow Attack trains flow-erasing \emph{perturbations} $\delta_t$, whose \ltwo\ \emph{norm} is controlled via $\varepsilon_2$.}
\label{fig:target_proximity_RAFT}
\end{figure}

Optical flow describes the apparent motion between subsequent frames of an image sequence.
It has numerous applications ranging from action recognition \cite{Ullah2019ActionReconition} and video processing \cite{Wang2020Superresolution} to robot navigation \cite{Zhang2020Robotics} and epidemic spread analysis~\cite{Stegmaier2020DifferencesEpidemicSpread}.
Over the past decade, the quality of optical flow methods has improved dramatically due to methodological advances: While early optical flow methods were mostly variational \cite{Horn1981HornSchunck,Black1993FrameworkRobustEstimation,Brox2004Warping,Brox2010LDOF,Sun2010Secrets}, today's top methods are based on neural networks~\cite{Ilg2017Flownet2Evolution,Ranjan2017OpticalFlowEstimation,Sun2018PwcNetCnns,Yang2019_VCN,Yin2019_HD3,teed2020raft,Jiang2021LearningEstimateHidden,Zhang2021_Separable}.

Up to date, theses methodological advances are mainly driven by quality scores on a few major benchmarks~\cite{Barron1994Performance,Baker2011Middlebury,Geiger2012AreWeReady,Butler2012NaturalisticOpenSource,Menze2015Joint3dEstimation}~that measure how well the calculated flow matches a known ground truth.
Given that optical flow is also used in the context of medical applications \cite{Yu2020CardiacMotion,Tehrani2020Ultrasound} and autonomous driving \cite{Capito2020OpticalFlowBased,Wang2021EndEndInteractive}, it is surprising that robustness plays only a subordinated role in the development of new methods.
In fact, robustness is rarely assessed in the literature and only few methods were developed with robustness as explicit goal \cite{Black1993FrameworkRobustEstimation,Weijer2004RobustOpticalFlow,Stein2004Census,Liu2010SIFTFlow,Li2018RobustOpticalFlow}.

A possible explanation for this blind spot is the ambiguity of the term \emph{robustness} as well as its challenging quantification for optical flow methods.
In this work we therefore focus on an improved measure for quantifying the robustness by means of adversarial attacks.
Our choice is motivated by the recently demonstrated vulnerability of optical flow networks to malicious input changes~\cite{Ranjan2019AttackingOpticalFlow}. 
This vulnerability clearly suggests that adversarial robustness should complement the qualitative performance when evaluating optical flow methods.

Adversarial attacks for optical flow are a very recent field of research with only two attacks available so far.
While Ranjan \etal~\cite{Ranjan2019AttackingOpticalFlow} proposed a local attack in terms of a patch-based approach, Schrodi \etal~\cite{Schrodi2022TowardsUnderstandingAdversarial} introduced a global attack inspired by attacks for classification.
This raises the question whether these two attacks are sufficiently strong to meaningfully measure adversarial robustness.
Answering it is difficult, as clear definitions for \emph{attack strength} and \emph{adversarial robustness} are currently missing in the context of optical flow.

In the context of classification, however, these quantities are already defined.
There, adversarial networks aim to find small perturbations to the input that lead to its misclassification \cite{Goodfellow2014ExplainingHarnessingAdversarial}.
Hence, stronger attacks need smaller input perturbations to cause an incorrect class.
While classification networks output a finite amount of discrete classes, optical flow methods predict a field of 2D flow vectors.
Using a small perturbation makes it unlikely that one can create an arbitrarily large deviation from the unperturbed flow.
Therefore, a sensible definition for a strong attack is \enquote{an attack that finds the most destructive adversarial perturbation from all perturbations under a specified bound}.

This subtle change in the definition of attack strength for optical flow significantly influences the attack design:
For a strong attack, an efficient way to bound the input perturbation is required; see Fig~\ref{fig:target_proximity_RAFT}.
Previous attacks for optical flow either lack effective bounds for their adversarial perturbations~\cite{Schrodi2022TowardsUnderstandingAdversarial}, or provide a weak adversarial attack~\cite{Ranjan2019AttackingOpticalFlow} to enable real-world applicability.
More effective attacks are therefore required for a rigorous quantification of adversarial robustness for optical flow.
In this context, we make the following contributions:
\begin{enumerate}
\item We formalize a generic \emph{threat model} for optical flow attacks and propose measures for \emph{attack strength} and \emph{adversarial robustness}, to improve the comparability of robustness evaluations and among adversarial attacks.
\item We present the \emph{Perturbation-Constrained Flow Attack (PCFA)}, a strong, global adversarial attack for optical flow that is able to limit the perturbation's \ltwo\ norm to remain within a chosen bound.
\item With PCFA, we generate \emph{joint} and \emph{universal} global perturbations.
\item We experimentally demonstrate that PCFA finds \emph{stronger adversarial samples} and is therefore better suited to quantify adversarial robustness than previous optical flow attacks.
\item We provide the first \emph{ranking} of current optical flow methods that combines their \emph{prediction quality} on benchmarks~\cite{Menze2015Joint3dEstimation,Butler2012NaturalisticOpenSource} with their \emph{adversarial robustness} measured by PCFA.
\end{enumerate}

%%%%%%%%%%%%%%%%%%%%%%%%%%%%%%%%%%%%%%%%%%%%%%%%%%%%%%%%%%%%%%%%%%%%%%%%%%%%%%%%%%%%%%%%%%%%%%%%%%%%%%%%%%%%
%%%%%%%%%%%%%%%%%%%%%%%%%%%%%%%%%%%%%%%%%%%%%%%%%%%%%%%%%%%%%%%%%%%%%%%%%%%%%%%%%%%%%%%%%%%%%%%%%%%%%%%%%%%%
%%%%
%%%%  RELATED WORK
%%%%
%%%%%%%%%%%%%%%%%%%%%%%%%%%%%%%%%%%%%%%%%%%%%%%%%%%%%%%%%%%%%%%%%%%%%%%%%%%%%%%%%%%%%%%%%%%%%%%%%%%%%%%%%%%%
%%%%%%%%%%%%%%%%%%%%%%%%%%%%%%%%%%%%%%%%%%%%%%%%%%%%%%%%%%%%%%%%%%%%%%%%%%%%%%%%%%%%%%%%%%%%%%%%%%%%%%%%%%%%

\section{Related Work}

In the following, we mainly focus on related work in the field of optical flow.
Thereby, we cover the assessment of robustness, the use of adversarial attacks as well as the design of neural networks. Further related work, also including adversarial attacks for classification, is discussed in our short review in Sec. \ref{sec:foundations}.

\medskip
\noindent\textbf{Robustness Assessment for Optical Flow.}
For assessing the robustness of optical flow methods, different concepts have been proposed in the literature.
On the one hand, early optical flow methods investigated robustness with regard to outliers~\cite{Black1993FrameworkRobustEstimation}, 
noise~\cite{Brox2004Warping,Bruhn2005CLG} or illumination changes \cite{Weijer2004RobustOpticalFlow}.
On the other hand, the Robust Vision Challenge\footnote{\href{http://www.robustvision.net/}{http://www.robustvision.net/}} quantifies robustness as the generalization of a method's qualitative performance across datasets. 
In contrast, our work relies on a different concept.
We consider \emph{adversarial robustness}~\cite{Szegedy2014IntriguingPropertiesNeural,Goodfellow2014ExplainingHarnessingAdversarial}, which is motivated by the Lipschitz continuity of functions. 
The Lipschitz constant is frequently used as a robustness measure for neural networks, where a small Lipschitz constant implies that the output does only change to the same extend as the input.
While finding the exact Lipschitz constant for a neural network is NP-hard~\cite{Virmaux2018LipschitzRegularityDeep}, bounding it is feasible.
For upper bounds on the Lipschitz constant, analytic architecture-dependent considerations are required that are generally difficult -- especially for such diverse architectures as in current flow networks.
In contrast, finding lower bounds is possible by performing adversarial attacks~\cite{Carlini2017TowardsEvaluatingRobustness}.
Hence, this work uses adversarial attacks to quantify robustness. Thereby, we aim to find input perturbations that cause particularly strong output changes.

\medskip
\noindent\textbf{Adversarial Attacks for Optical Flow.}
To the best of our knowledge, there are only two works that propose adversarial attacks tailored to optical flow networks so far.
Ranjan \etal~\cite{Ranjan2019AttackingOpticalFlow} developed the first adversarial attack, which causes wrong flow predictions by placing a colorful circular patch in both input frames.
To be applicable in the wild, their patches are trained with many constraints (e.g.\ location and rotation invariances, patches are circular coherent regions), which comes at the cost of a reduced attack strength.
Recently, Schrodi \etal~\cite{Schrodi2022TowardsUnderstandingAdversarial} introduced a less constrained global attack on optical flow, based on the I-FGSM~\cite{Kurakin2017AdversarialMachineLearning} attack for classification.
While posing fewer constraints than the Patch Attack, FGSM~\cite{Goodfellow2014ExplainingHarnessingAdversarial} and the iterative I-FGSM attacks were developed for speed rather than attack strength and furthermore do not effectively limit the perturbation size\footnote{FGSM \cite{Goodfellow2014ExplainingHarnessingAdversarial} and I-FGSM\cite{Kurakin2017AdversarialMachineLearning} limit the perturbation size below $\varepsilon_\infty$ by performing only so many steps of a fixed step size $\tau$, that exceeding the norm bound is impossible. To this end,
the number of steps is fixed to $N=\lfloor \frac{\varepsilon_\infty}{\tau} \rfloor$, which comes down to a one-shot optimization. Additionally, this \enquote{early stopping} reduces the attack strength, for it prevents optimizing in the vicinity of the bound.}.
Hence, none of the current flow attacks is fully suitable for a rigorous adversarial robustness assessment. In contrast, our novel PCFA method does not make compromises regarding the attack strength, while effectively allowing to keep the perturbations below a specified bound.

In the context of vision problems similar to optical flow, Wong \etal~\cite{Wong2021StereopagnosiaFoolingStereo} successfully attacked stereo networks with I-FGSM~\cite{Kurakin2017AdversarialMachineLearning} and its momentum variant MI-FGSM~\cite{Dong2018BoostingAdversarialAttacks}.
Moreover, Anand \etal~\cite{Anand2020AdversarialPatchDefense} proposed an approach to secure optical flow networks for action recognition against adversarial patch attacks by a preceding filtering step that detects, removes and inpaints the attacked location.

\medskip
\noindent\textbf{Neural Networks for Optical Flow.}
Regarding neural networks for optical flow, related work is given by those approaches for which we later on evaluate the robustness, i.e.\ the methods in \cite{Ilg2017Flownet2Evolution,Ranjan2017OpticalFlowEstimation,Sun2018PwcNetCnns,teed2020raft,Jiang2021LearningEstimateHidden}. 
These approaches are representatives of the following three classes of networks: classical, pyramidal and recurrent networks.
\emph{Classical networks} such as FlowNet2~\cite{Ilg2017Flownet2Evolution} rely on a stacked encoder-decoder architecture with a dedicated feature extractor and a subsequent correlation layer. 
More advanced \emph{pyramidal networks} such as SpyNet~\cite{Ranjan2017OpticalFlowEstimation} and PWCNet~\cite{Sun2018PwcNetCnns} estimate the optical flow in coarse-to-fine manner using initializations from coarser levels, by warping either the input frames or the extracted features.
Finally, state-of-the-art \emph{recurrent networks} such as RAFT~\cite{teed2020raft} and GMA~\cite{Jiang2021LearningEstimateHidden} perform iterative updates that rely on a sampling-based hierarchical cost volume.
Thereby GMA additionally considers globally aggregated motion features to improve the performance at occlusions.

\section{Adversarial Attacks: Foundations and Notations}
\label{sec:foundations}

Adversarial attacks uncovered the brittle performance of neural networks on slightly modified input images, so called \emph{adversarial samples}~\cite{Szegedy2014IntriguingPropertiesNeural,Goodfellow2014ExplainingHarnessingAdversarial}.
Different ways to train adversarial samples exist, which lead to different attack types.
\emph{Targeted attacks} perturb the input to induce a specified target output. Compared to \emph{untargeted attacks}, they are considered the better choice for strong attacks, since they can simulate the former ones by running attacks on all possible targets and taking the most successful perturbation~\cite{Carlini2017TowardsEvaluatingRobustness}.
\emph{Global attacks}~\cite{Goodfellow2014ExplainingHarnessingAdversarial,Carlini2017TowardsEvaluatingRobustness,Kurakin2017AdversarialMachineLearning,Dong2018BoostingAdversarialAttacks} allow any pixel of the image to be disturbed within a norm bound, while \emph{patch attacks}~\cite{Brown2018AdversarialPatch,Ranjan2019AttackingOpticalFlow} perturb only pixels within a certain neighborhood.
So called \emph{universal perturbations} are particularly transferable and have a degrading effect on multiple images rather than being optimized for a single one~\cite{MoosaviDezfooli2017UniversalAdversarialPerturbations,Shafahi2020UniversalAdversarialTraining,Deng2020UniversalAdversarialAttack,Ranjan2019AttackingOpticalFlow,Schrodi2022TowardsUnderstandingAdversarial}.
As they affect a class of input images, their effect on a single image is often weaker.

Attacks for optical flow followed five years after the existence of adversarial samples was discovered for classification networks~\cite{Szegedy2014IntriguingPropertiesNeural,Goodfellow2014ExplainingHarnessingAdversarial}.
As concepts from classification attacks are present in all flow attacks, we briefly review those attacks before we discuss attacks for optical flow in more detail.

\medskip
\noindent\textbf{Classification.}
Szegedy \etal~\cite{Szegedy2014IntriguingPropertiesNeural} provided the first optimization formulation for a targeted adversarial attack to find a small perturbation $\delta \in \Rds^m$ to the input $x\in \Rds^m$, such that a classifier $\mathcal{C}$ outputs the incorrect label $t$:
\begin{equation}
\label{equ:adv_att_optim_class}
\min \|\delta\|_2\ \quad \text{s.t.} \quad \mathcal{C}(x+\delta) = t\, , \quad \text{and}\quad x+\delta \in [0,1]^m \, .
\end{equation}
Many adversarial attacks were proposed to solve problem~\eqref{equ:adv_att_optim_class}, with varying focus and applicability~\cite{Goodfellow2014ExplainingHarnessingAdversarial,Kurakin2017AdversarialMachineLearning,Dong2018BoostingAdversarialAttacks,Carlini2017TowardsEvaluatingRobustness,Brown2018AdversarialPatch}.
In the following we discuss the two methods that are most relevant to our work.
On the one hand, there is the Fast Gradient Sign Method (FGSM)~\cite{Goodfellow2014ExplainingHarnessingAdversarial} method which is used for a fast generation of adversarial samples, with later variations including multiple iterations~\cite{Kurakin2017AdversarialMachineLearning} and momentum~\cite{Dong2018BoostingAdversarialAttacks}.
On the other hand, there is the C\&W attack~\cite{Carlini2017TowardsEvaluatingRobustness} which emphasizes perturbation destructivity by encouraging the target over all other labels in the optimization~\eqref{equ:adv_att_optim_class}, while minimizing the adversarial perturbation's \ltwo\ norm. 
For a broader overview of attacks, we refer to the review article of Xu \etal~\cite{Xu2020AdversarialAttacksDefenses}.

\medskip
\noindent\textbf{Optical Flow.}
Optical flow describes the apparent motion of pixels between two subsequent frames $\img_t$ and $\img_{t+1} \in \Rds^{C \times I}$ of an image sequence in terms of a displacement vector field over the image area $I \!=\! M\! \times \!N$.
This yields the ground truth flow field $\fgt = (u^g,v^g) \in \Rds^{2 \times I}$.
From the initial frames  $\img_t,\, \img_{t+1} \in [0,1]^{C\times I}$ with $C$ channels each, a flow network predicts the \emph{unattacked} or \emph{initial optical flow} $\flow \in \Rds^{2 \times I}$.
Adding adversarial perturbations $\delta_t, \delta_{t+1} \in \Rds^{C\times I}$ to the inputs yields the \emph{perturbed} or \emph{adversarial flow} $\fadv \in \Rds^{2 \times I}$.

The \emph{local} Patch Attack by Ranjan \etal~\cite{Ranjan2019AttackingOpticalFlow} optimizes universal patches
\begin{equation}
\underset{\delta}{\text{argmin}}\  \loss(\fadv, \targ) \quad \text{s.t.} \quad \delta_t = \delta_{t+1} = \delta\, , \quad \delta \text{ is patch} \quad \text{and}\quad \delta \in [0,1]^{C\times I}
\end{equation}
over multiple frames.
The cosine similarity serves as loss function $\loss$ to minimize the angle between $\fadv$ and $\targ$, targeting the negative initial flow $\targ = - \flow$.
To make the circular perturbations location and rotation invariant, the respective transformations are applied before adding the perturbation to the frames.

In contrast, the \emph{global} flow attack by
Schrodi \etal~\cite{Schrodi2022TowardsUnderstandingAdversarial} uses $N$ steps of I-FGSM~\cite{Kurakin2017AdversarialMachineLearning} to generate adversarial perturbations $\|\delta_t, \delta_{t+1}\|_\infty \leq \varepsilon_\infty$ for single frames as
\begin{equation}
\label{equ:FGSM}
\delta_z^{(n+1)} = \delta_z^{(n)} - \frac{\varepsilon_\infty}{N} \cdot \text{sgn}(\nabla_{\img_z+\delta_z^{(n)}} \loss(\fadv,\targ))\, , \quad z = t,\, t+1\, , \ \ n = 1, \hdots, N .
\end{equation}
For universal perturbations, it uses the loss function and training from~\cite{Ranjan2019AttackingOpticalFlow}.

While we also develop a global attack like~\cite{Schrodi2022TowardsUnderstandingAdversarial}, we do not base it on I-FGSM variants as they quickly generate non-optimal perturbations.
Instead, we develop a novel attack for optical flow, and explicitly optimize it for attack strength.
An inspiration we take from classification methods is a way to keep perturbed frames in a valid color range, which was proposed for the C\&W method~\cite{Carlini2017TowardsEvaluatingRobustness}.

%%%%%%%%%%%%%%%%%%%%%%%%%%%%%%%%%%%%%%%%%%%%%%%%%%%%%%%%%%%%%%%%%%%%%%%%%%%%%%%%%%%%%%%%%%%%%%%%%%%%%%%%%%%%
%%%%%%%%%%%%%%%%%%%%%%%%%%%%%%%%%%%%%%%%%%%%%%%%%%%%%%%%%%%%%%%%%%%%%%%%%%%%%%%%%%%%%%%%%%%%%%%%%%%%%%%%%%%%
%%%%
%%%%  METHODS
%%%%
%%%%%%%%%%%%%%%%%%%%%%%%%%%%%%%%%%%%%%%%%%%%%%%%%%%%%%%%%%%%%%%%%%%%%%%%%%%%%%%%%%%%%%%%%%%%%%%%%%%%%%%%%%%%
%%%%%%%%%%%%%%%%%%%%%%%%%%%%%%%%%%%%%%%%%%%%%%%%%%%%%%%%%%%%%%%%%%%%%%%%%%%%%%%%%%%%%%%%%%%%%%%%%%%%%%%%%%%%

\section{A Global Perturbation-Constrained Adversarial Attack for Optical Flow Networks}

As motivated in the introduction, strong flow attacks require refined notions for \emph{attack strength} and \emph{adversarial robustness}, which we discuss first.
Based on these refined notions, we present the Perturbation-Constrained Flow Attack (PCFA), which optimizes for strong adversarial perturbations while keeping their \ltwo\ norm under a specified bound.
Moreover, we discuss different perturbation types for optical flow attacks such as joint and universal perturbations.

\subsection{Attack Strength and Adversarial Robustness for Optical Flow}

In the context of classification networks, a strong adversarial sample is one that causes a misclassification while being small, see Problem~\eqref{equ:adv_att_optim_class}.
As optical flow methods do not produce discrete classes but flow fields $\flow \in \Rds^{2\times I}$, significantly larger adversarial perturbations $\delta_t, \delta_{t+1}$ would be required to induce a specific target flow $\targ$.
Further it is unclear whether a method \emph{can} output a certain target.
What can be controlled, however, is the perturbation size.
Therefore, a useful \emph{threat model for optical flow} is one that limits the perturbation size to $\varepsilon$ and minimizes the distance between attacked flow and target at the same time:
\begin{equation}
\label{equ:minimization_constr}
\underset{\delta_t, \delta_{t+1}}{\text{argmin}}\ \loss(\fadv,\targ) \ \  \text{s.t.} \ \  \|\delta_t, \delta_{t+1}\| \leq \varepsilon\, , \ \ \text{and}\ \ \img_z+\delta_z\ \in [0,1]^{C\times I}\, , \ \  z=t,\, t+1\, .
\end{equation}
Because the used norms and bounds are generic in this formulation, previous flow attacks fit into this framework: 
The Patch Attack by Ranjan \etal~\cite{Ranjan2019AttackingOpticalFlow} poses a \lzero\ bound on the patch by limiting the patch size, while the I-FGSM flow attack by Schrodi \etal~\cite{Schrodi2022TowardsUnderstandingAdversarial} can be seen as a \linfty\ bound on the perturbation.

\medskip
\noindent\textbf{Quantifying Attack Strength.}
Given that two attacks use identical targets, norms and bounds, they are fully comparable in terms of effectiveness.
For stronger attacks, the adversarial flow $\fadv$ resembles the target $\targ$ better.
As the average endpoint error AEE (see Eq.~\eqref{equ:AEE}) is widely used to quantify distances between optical flow fields, we propose the following quantification of \emph{attack strength} as
\begin{center}
AEE$(\fadv,\targ) \ \ \text{for} \ \ \|\delta_t, \delta_{t+1}\| \leq \varepsilon \quad=\quad$Attack Strength.
\end{center}

\medskip
\noindent\textbf{Quantifying Adversarial Robustness.}
To assess the general robustness of a given method, the specific target is of minor importance.
A robust method should not produce a largely different flow prediction for slightly changed input frames, which makes the distance between adversarial flow $\fadv$ and unattacked flow $\flow$ a meaningful metric.
This distance is small for robust methods and can be quantified with the AEE.
Therefore we propose to quantify the \emph{adversarial robustness} for optical flow as
\begin{center}
AEE$(\fadv,\flow) \ \ \text{for} \ \ \|\delta_t, \delta_{t+1}\| \leq \varepsilon \quad=\quad$Adversarial Robustness.
\end{center}

This definition of adversarial robustness does intentionally not include a comparison to the ground truth flow $\fgt$ as in~\cite{Ranjan2017OpticalFlowEstimation, Schrodi2022TowardsUnderstandingAdversarial} for two reasons.
First, a ground truth com\-parison measures the flow quality, which should be kept separate from robustness because these quantities likely hinder each other~\cite{Tsipras2019RobustnessMayBe}.
Secondly, changed frame pairs will have another ground truth, but it is unclear what this ground truth for the attacked frame pairs looks like.
While constructing a pseudo ground truth for patch attacks can be possible\footnote{Ranjan \etal~\cite{Ranjan2019AttackingOpticalFlow} generate a pseudo ground truth for their attack with static patches by prescribing a zero-flow at the patch locations.}, the required ground truth modifications for global attacks are in general unknown.
For those reasons, we suggest to report quality metrics and adversarial robustness separately.

\subsection{The Perturbation-Constrained Flow Attack}

Starting with the threat model for optical flow~\eqref{equ:minimization_constr}, we opt for global perturbations to generate strong adversarial samples.
To obtain a differentiable formulation, we bound the \ltwo\ norm of the perturbation.
Our \emph{Perturbation-Constrained Flow Attack (PCFA)} then solves the inequality-constrained optimization
\begin{equation}
\label{equ:minimization_PCFA}
\underset{\delta_t, \delta_{t+1}}{\text{argmin}}\ \loss(\fadv,\targ) \ \ \ \text{s.t.} \ \ \ \|\delta_t, \delta_{t+1}\|_2 \leq \varepsilon_2 \sqrt{2IC} , \quad \img_z\!+\!\delta_z\!\in\![0,1]^{C\times I}\! ,\ \   z=t, t\!+\!1 .
\end{equation}
We use the additional factor $\sqrt{2IC}$ to make the perturbation bound independent of the image size $I=M\times N$ and channels $C$.
This way, $\varepsilon_2 = 0.01$ signifies an average distortion of 1\% of the frames' color range per pixel.
To solve \eqref{equ:minimization_PCFA}, four aspects need consideration:
(i) How to implement the inequality constraint $\|\delta_t, \delta_{t+1}\|_2 \leq \varepsilon \sqrt{2IC}$, (ii) how to choose the loss function $\loss$, (iii) how to choose the target $\targ$ and (iv) how to ensure the box constraint $\img_z+\delta_z\ \in [0,1]^{C\times I}$, $z = t,\,t\!+\!1$.

\medskip
\noindent\textbf{Inequality Constraint.}
We use a penalty method with exact penalty function~\cite{Nocedal2006NumericalOptimization} to transform the inequality-constrained problem~\eqref{equ:minimization_PCFA} into the following unconstrained optimization problem for $\hat{\delta} = \delta_t, \delta_{t+1}$:
\begin{equation}
\label{equ:minimization_unconstr}
\underset{\hat{\delta}}{\text{argmin}}\  \phi(\hat{\delta}, \mu)\, , \quad \phi(\hat{\delta}, \mu) = \mathcal{L}(\fadv,\targ) + \mu | c(\hat{\delta})|\, .
\end{equation}
The penalty function $c$ linearly penalizes deviations from the constraint $\|\hat{\delta}\|_2 \leq \hat{\varepsilon}_2 = \varepsilon_2 \sqrt{2IC}$ and is otherwise zero: $c(\hat{\delta}) = \max(0, \|\hat{\delta}\|_2 - \hat{\varepsilon}_2) = \text{ReLU}(\|\hat{\delta}\|_2 - \hat{\varepsilon}_2)$.
If the penalty parameter $\mu \in \Rds$ approaches infinity, the unconstrained problem~\eqref{equ:minimization_unconstr} will take its minimum within the specified constraint.
In practice, it is sufficient to choose $\mu$ large.
The selected exact penalty function $\phi$ is nonsmooth at $\|\hat{\delta}\|_2 = \hat{\varepsilon}_2$, which makes its optimization potentially problematic.
However, formulating the problem with a smooth penalty function would require to solve a series of optimization problems and is therefore computationally expensive~\cite{Nocedal2006NumericalOptimization}.
We solve~\eqref{equ:minimization_unconstr} directly with the L-BFGS~\cite{Nocedal1980UpdatingQuasiNewton} optimizer, which worked well in practice.
Moreover, in our implementation we use the squared quantities $\|\hat{\delta}\|_2^2 \leq \hat{\varepsilon}_2^2$ for the constraint to avoid a pole in the derivative of $\|\hat{\delta}\|_2$ at $\hat{\delta}=0$.

\medskip
\noindent\textbf{Loss Functions.}
The loss function $\loss$ should quantify the proximity of adversarial and target flow.
The \emph{average endpoint error (AEE)} is a classical measure to quantify the quality of optical flow as
\begin{equation}
\label{equ:AEE}
\text{AEE}(\fadv, \targ) = \frac{1}{I} \sum_{i \in I} \|\fadv_{i} - \targ_{i}\|_2 \, .
\end{equation}
However, its derivative is undefined if single components of the adversarial- and target flow coincide, i.e.\ $\fadv_{j} = \targ_{j}$.
In practice, we rarely observed problems for reasonably small perturbations bounds $\hat{\varepsilon}_2$, which prevent a perfect matching. 
The \emph{mean squared error (MSE)} circumvents this issue due to its squared norm
\begin{equation}
\text{MSE}(\fadv, \targ) = \frac{1}{I} \sum_{i\in I}  \|\fadv_{i} - \targ_{i}\|_2^2 \, ,
\end{equation}
but is less robust to outliers as deviations are penalized quadratically.
Previous optical flow attacks~\cite{Ranjan2019AttackingOpticalFlow,Schrodi2022TowardsUnderstandingAdversarial} use the \emph{cosine similarity (CS)}
\begin{equation}
\label{equ:cossim}
\text{CS}(\fadv, \targ) = \frac{1}{I} \sum_{j=1}^{I}  \frac{\langle \fadv_{j}, \targ_{j} \rangle}{\|\fadv_{j}\|_2 \cdot \|\targ_{j}\|_2}\, .
\end{equation}
Since this loss only measures angular deviations between flows, it fails to train adversarial perturbation where the adversarial flow or the target are zero.

\medskip
\noindent\textbf{Target Flows.}
In principle, any flow field can serve as target flow.
Ranjan \etal~\cite{Ranjan2019AttackingOpticalFlow} flip the flow direction for a \emph{negative-flow attack} $\targ = -\flow$.
However, this target strongly depends on the initial flow direction.
As input agnostic alternative, we propose the \emph{zero-flow attack} with $\targ = 0$.
It is especially useful to train universal perturbations that are effective on multiple frames.

\medskip
\noindent\textbf{Ensuring the Box Constraint.}
During the optimization, the perturbed frames should remain within the allowed value range, e.g.\ return a valid color value.
All previous flow attacks use \emph{clipping} that crops the perturbed frames to their allowed range after adding $\delta_t, \delta_{t+1}$.
The \emph{change of variables (COV)}~\cite{Carlini2017TowardsEvaluatingRobustness} is an alternative approach that optimizes over the auxiliary variables $w_t, w_{t+1}$ instead of $\delta_t, \delta_{t+1}$ as
\begin{equation}
\delta_z = \frac{1}{2}(\text{tanh}(w_z) + 1) - \mathcal{I}_z\, , \quad z = t,\, t+1\, .
\end{equation}
This optimizes $w_t, w_{t+1} \in [-\infty, \infty]$, and afterwards maps them into the allowed range $[0,1]$ of the perturbed frames.
Our evaluation considers both approaches.

\subsection{Joint and Universal Adversarial Perturbations}

\begin{figure}[tb]
\centering
   \small
   \centering
   \begin{tabular}{@{}r@{\ \ \ \ \ \ }@{\ }M{70mm}@{\quad\quad}M{70mm}@{}}
            & \textbf{Frame-Specific} & \textbf{Universal}  \\[6pt]
   \textbf{Disjoint} & \includegraphics[width=70mm]{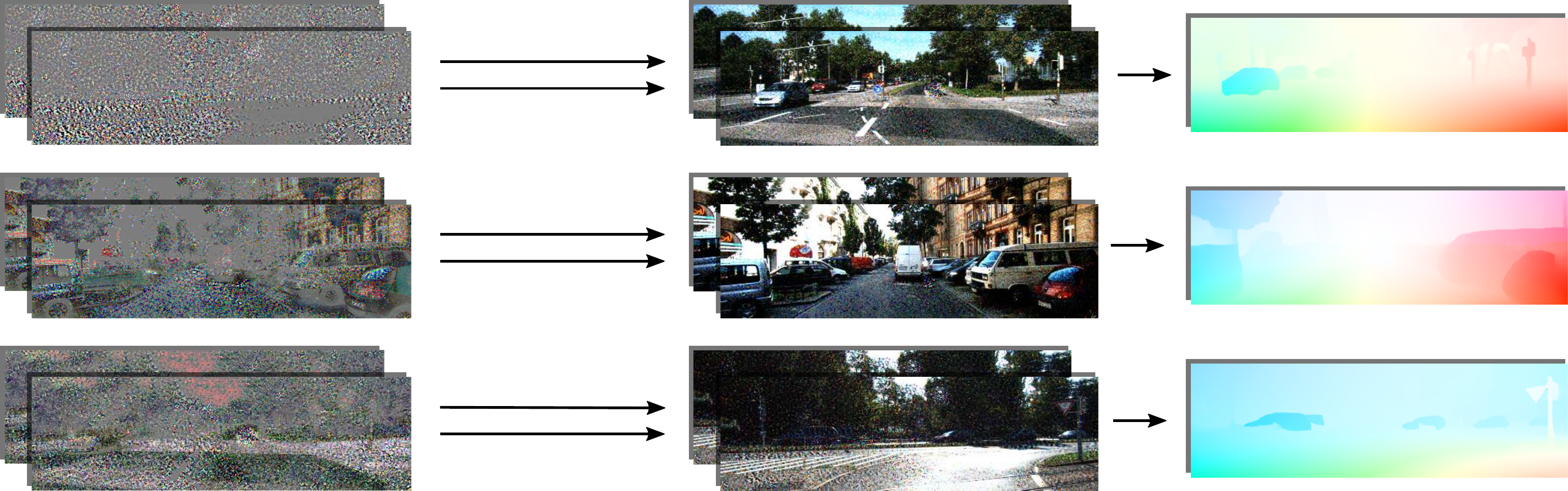} & \includegraphics[width=70mm]{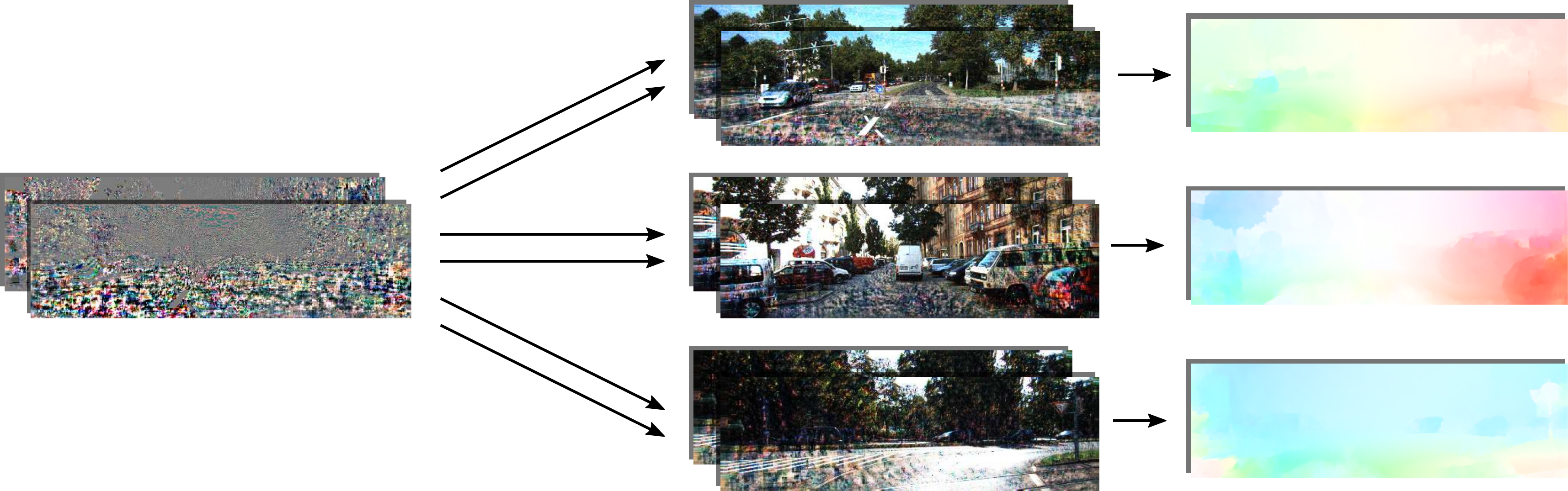} 
   \\[-3pt]
            & \multicolumn{1}{l}{\phantom{ll}$\dd$} & \multicolumn{1}{l}{\phantom{al}$\ddu$}  \\[3pt]
   \textbf{Joint}   & \includegraphics[width=70mm]{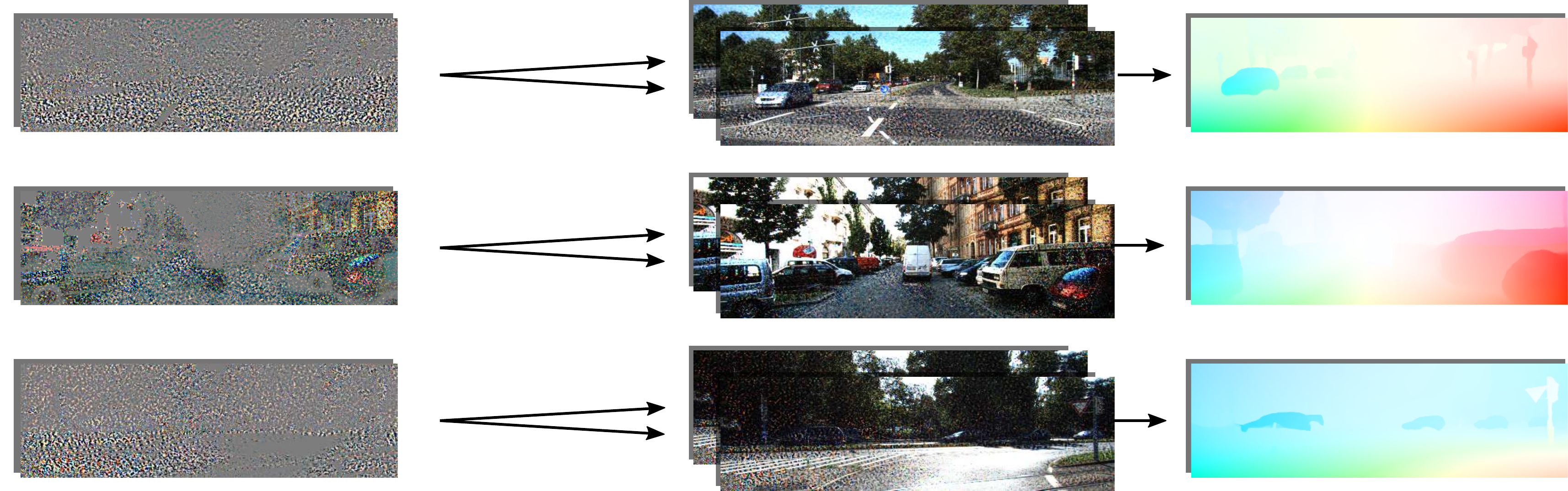} & \includegraphics[width=70mm]{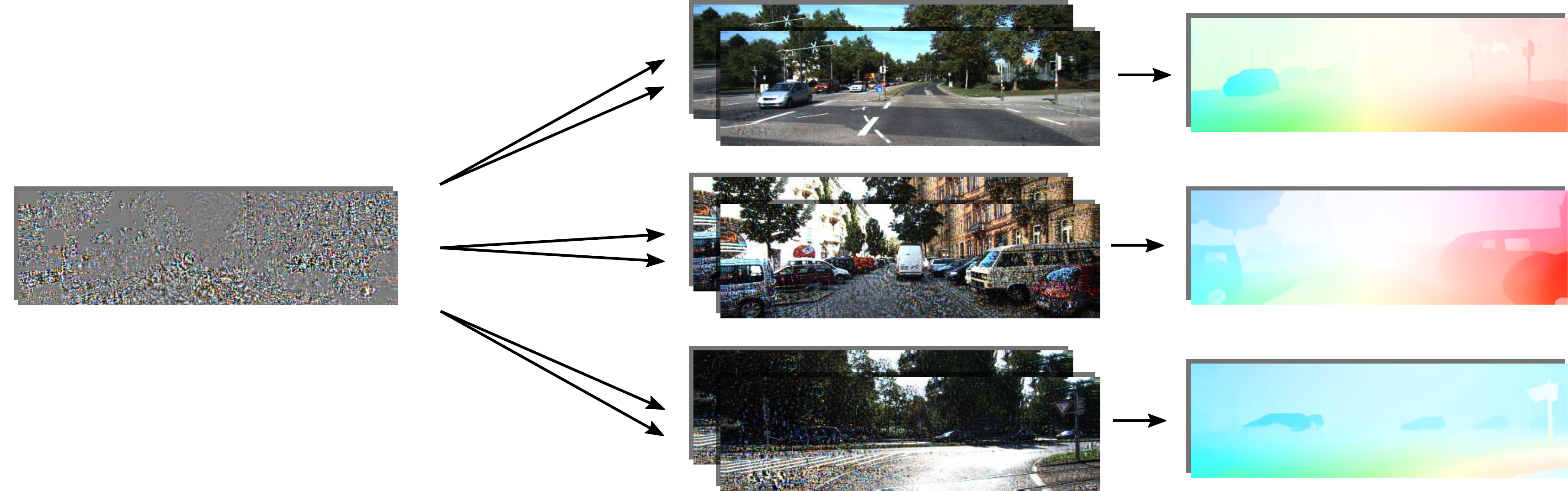} 
   \\[-4pt]
            & \multicolumn{1}{l}{\phantom{al}$\dc$} & \multicolumn{1}{l}{\phantom{lm}$\dcu$}  \\
   \end{tabular}
\caption{Illustration of the differences between \emph{disjoint} and \emph{joint} as well as \emph{frame-specific} and \emph{universal} adversarial perturbations for attacking optical flow networks.}
\label{fig:dd_dc_ddu_dcu}
\end{figure}

Out of the box, the optimization problem~\eqref{equ:minimization_unconstr} holds for \emph{disjoint} perturbations, resulting in two perturbations $\dd$ for an input frame pair $\img_t,\, \img_{t+1}$.
Let us now discuss optimizing \emph{joint perturbations} for both input frames and \emph{universal perturbations} for multiple input pairs; their difference is illustrated in Fig.~\ref{fig:dd_dc_ddu_dcu}.
By attacking both frames or frame pairs simultaneously, these perturbations 
have to fulfill more constraints and are therefore 
expected to offer a weaker performance.

\medskip
\noindent\textbf{Joint Adversarial Perturbations.}
In case of \emph{joint} adversarial perturbations, a common perturbation $\dc$ is added to both input frames.
In its current formulation, the COV box constraint is only possible for disjoint perturbations.

\medskip
\noindent\textbf{Universal Adversarial Perturbations.}
Training \emph{universal} instead of \emph{frame-specific} perturbations is straightforward using our optimization~\eqref{equ:minimization_unconstr}.
Additional projection operations to ensure the norm bound as in other schemes~\cite{MoosaviDezfooli2017UniversalAdversarialPerturbations,Deng2020UniversalAdversarialAttack,Shafahi2020UniversalAdversarialTraining,Ranjan2019AttackingOpticalFlow,Schrodi2022TowardsUnderstandingAdversarial} are unnecessary, because PCFA directly optimizes perturbations of limited size.
Similar to~\cite{Shafahi2020UniversalAdversarialTraining}, we refine adversarial perturbations on minibatches.
With this scheme, we train disjoint $\ddu$ and joint $\dcu$ universal perturbations.

\subsection{Design Overview and Comparison to Literature}
Tab.~\ref{table:attack_comparison} summarizes our method design-wise and compares it to the other optical flow attacks from the literature.
PCFA is the first attack that allows an effective \ltwo\ norm bound for the perturbation. 
In the evaluation, we provide an extensive analysis of its performance for different perturbation types, losses and targets.

\setlength{\tabcolsep}{4pt}
\begin{table}
\begin{center}
\caption{Comparison of adversarial optical flow attacks, configurations as stated in the respective publications. \emph{Clip} = Clipping, 
\emph{A} = AAE, \emph{M} = MSE, \emph{C} = CS.}
\label{table:attack_comparison}
\scalebox{1}{
\begin{tabular}{@{\ }lllccccccc@{\ }}
\toprule
\multirow{2}{*}{Attack} & \multirow{2}{*}{Type} &  \multirow{2}{*}{$\|\delta\|_*$} & 
\multicolumn{4}{c}{Perturbation Types}&
\multirow{2}{*}{Losses} & \multirow{2}{*}{Box Constr.}\\
\cmidrule(l{0.3em}r{0.3em}){4-7}
                        &                       &                                  & $\dd$ & $\dc$ & $\ddu$ & $\dcu$ \\
\midrule
\textbf{Patch Att.}~\cite{Ranjan2019AttackingOpticalFlow} & Patch & \lzero & -- & -- & -- & \yes & C & Clip\\
\textbf{I-FGSM}~\cite{Goodfellow2014ExplainingHarnessingAdversarial,Schrodi2022TowardsUnderstandingAdversarial} & Global & \linfty & \yes & -- & \yes & -- & C & Clip\\
\textbf{PCFA} (ours) & Global &  \ltwo & \yes & \yes & \yes & \yes & A, M, C & Clip, COV\\
\bottomrule
\end{tabular}
}
\end{center}
\end{table}
\setlength{\tabcolsep}{1.4pt}

%%%%%%%%%%%%%%%%%%%%%%%%%%%%%%%%%%%%%%%%%%%%%%%%%%%%%%%%%%%%%%%%%%%%%%%%%%%%%%%%%%%%%%%%%%%%%%%%%%%%%%%%%%%%
%%%%%%%%%%%%%%%%%%%%%%%%%%%%%%%%%%%%%%%%%%%%%%%%%%%%%%%%%%%%%%%%%%%%%%%%%%%%%%%%%%%%%%%%%%%%%%%%%%%%%%%%%%%%
%%%%
%%%%  RESULTS / EXPERIMENTS
%%%%
%%%%%%%%%%%%%%%%%%%%%%%%%%%%%%%%%%%%%%%%%%%%%%%%%%%%%%%%%%%%%%%%%%%%%%%%%%%%%%%%%%%%%%%%%%%%%%%%%%%%%%%%%%%%
%%%%%%%%%%%%%%%%%%%%%%%%%%%%%%%%%%%%%%%%%%%%%%%%%%%%%%%%%%%%%%%%%%%%%%%%%%%%%%%%%%%%%%%%%%%%%%%%%%%%%%%%%%%%

\section{Experiments}
\label{sec:Experiments}

Our evaluation addresses three distinct aspects: 
First, we identify loss functions, box constraints and targets for PCFA that yield the strongest attack, and compare the resulting approach to I-FGSM~\cite{Schrodi2022TowardsUnderstandingAdversarial}.
Secondly, we assess the strength of PCFA's joint and universal perturbations in white- and black box attacks, which includes a comparison with the Patch Attack from \cite{Ranjan2017OpticalFlowEstimation}.
Finally, based on PCFA's strongest configuration, we perform a common evaluation of optical flow methods regarding estimation quality \emph{and} adversarial robustness.

At \href{https://github.com/cv-stuttgart/PCFA}{https://github.com/cv-stuttgart/PCFA} we provide our PCFA implementation in PyTorch~\cite{Paszke2019Pytorch}.
It is evaluated it with implementations of FlowNet2~\cite{Ilg2017Flownet2Evolution} from \cite{Reda2017Flownet2PytorchPytorch}, PWCNet~\cite{Sun2018PwcNetCnns}, SpyNet~\cite{Ranjan2017OpticalFlowEstimation} from~\cite{Niklaus2018PytorchSpyNet}, RAFT~\cite{teed2020raft} and GMA~\cite{Jiang2021LearningEstimateHidden} on the datasets KITTI 2015~\cite{Menze2015Joint3dEstimation} and MPI-Sintel final~\cite{Butler2012NaturalisticOpenSource}.
A full list of parameters and configurations for all experiments is in the supplementary material, Tab.~\ref{table:exp_configs}.

\subsection{Generating Strong Perturbations for Individual Frame Pairs}

In the following we consider \emph{disjoint non-universal} perturbations $\dd$ on the KITTI test dataset. 
This allows us to (i) identify the strongest PCFA configuration, to (ii) show that PCFA can be used to target specific flows, and to (iii) compare its strength to I-FGSM~\cite{Schrodi2022TowardsUnderstandingAdversarial}.
We solve PCFA from Eq.~\eqref{equ:minimization_unconstr} with $20$ L-BFGS~\cite{Nocedal1980UpdatingQuasiNewton} steps per frame pair.

\setlength{\tabcolsep}{4pt}
\begin{table}[b]
\begin{center}
\caption{PCFA attack strength $\text{AEE}(\fadv,\targ)$ on the KITTI test dataset for different \emph{loss functions}, \emph{targets} and \emph{box constraints} on RAFT. Small values indicate strong attacks.}
\label{table:setup_PCFA_loss_tgt_pt}
\begin{tabular}{@{}r@{\quad\quad}rrrcrrr@{}}
\toprule
          &  \multicolumn{3}{c}{$\targ = 0$}         & & \multicolumn{3}{c}{$\targ = -\flow$}                \\
          \cmidrule(l{0.em}r{0em}){2-4}  \cmidrule(l{0.2em}r{0em}){6-8}
         &  \multicolumn{1}{r}{AEE} 
         &  \multicolumn{1}{r}{MSE} 
         &  \multicolumn{1}{r}{CS} 
         &  \;\;\;\;
         &  \multicolumn{1}{r}{AEE} 
         &  \multicolumn{1}{r}{MSE} 
         &  \multicolumn{1}{r}{CS} \\
\midrule
Clipping &      
\phantom{00}3.76 &     
\phantom{0}10.51 &       
\phantom{0}32.37 & &       
\phantom{0}22.48 &       
\phantom{0}38.57 &           
129.44 \\
COV &      
\textbf{3.54} &      
\phantom{00}7.46 &        
\phantom{0}32.37 & &        
\textbf{18.84} &         
\phantom{0}34.82 &            
\phantom{0}86.00 \\
\bottomrule
\end{tabular}
\end{center}
\end{table}
\setlength{\tabcolsep}{1.4pt}

\medskip
\noindent\textbf{Loss and Box Constraint.}
Tab.~\ref{table:setup_PCFA_loss_tgt_pt} summarizes the attack strength $\text{AEE}(\fadv,\targ)$ for all combinations of losses and box constraints on the targets $\targ \in\{0,-\flow\}$ with $\varepsilon_2=5\cdot10^{-3}$ for RAFT. 
Using the same loss, change of variables (COV)\linebreak always yields a stronger attack than clipping, i.e.\ a smaller distance to the target.
Despite its problematic derivative, the average endpoint error (AEE) reliably outperforms the other losses, while the cosine similarity (CS) that is used in all previous flow attacks~\cite{Ranjan2019AttackingOpticalFlow,Schrodi2022TowardsUnderstandingAdversarial} performs worst.
Also, the CS loss fails on the zero-flow target where perturbations keep their initial values (\cf Supp.\ Tab.~\ref{table:setup_PCFA_loss_tgt_pp}).

Since AEE with COV yields the strongest attack independent of the target, we select this configuration for the remaining experiments.

\begin{figure}[tb]
\footnotesize
\centering
\setlength{\fboxrule}{0.1pt}%
\setlength{\fboxsep}{0pt}%
\begin{tabular}{@{}r@{\ }M{36mm}@{\ \ }M{36mm}@{\ }M{36mm}@{\ }M{36mm}@{}}
            & \footnotesize Initial Flow & \footnotesize $\varepsilon_2=5\cdot 10^{-4}$ & \footnotesize $\varepsilon_2=5\cdot 10^{-3}$ & \footnotesize $\varepsilon_2=5\cdot 10^{-2}$ \\
   \footnotesize FlowNet2 & \fcolorbox{gray}{white}{\includegraphics[width=36mm]{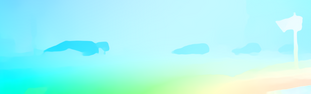}} & \fcolorbox{gray}{white}{\includegraphics[width=36mm]{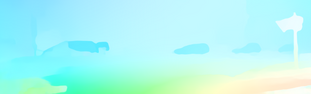}} & \fcolorbox{gray}{white}{\includegraphics[width=36mm]{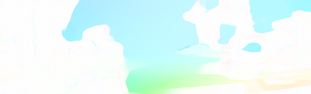}} & \fcolorbox{gray}{white}{\includegraphics[width=36mm]{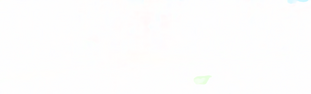}}  \\
   \footnotesize SpyNet & \fcolorbox{gray}{white}{\includegraphics[width=36mm]{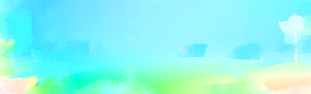}} & \fcolorbox{gray}{white}{\includegraphics[width=36mm]{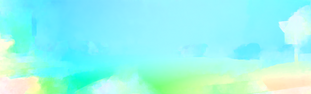}} & \fcolorbox{gray}{white}{\includegraphics[width=36mm]{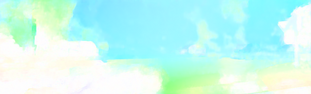}} & \fcolorbox{gray}{white}{\includegraphics[width=36mm]{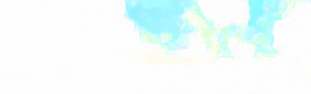}}  \\
   \footnotesize RAFT & \fcolorbox{gray}{white}{\includegraphics[width=36mm]{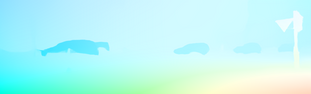}} & \fcolorbox{gray}{white}{\includegraphics[width=36mm]{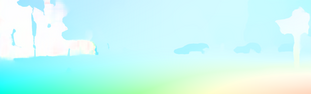}} & \fcolorbox{gray}{white}{\includegraphics[width=36mm]{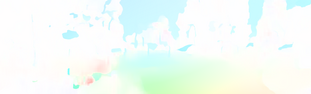}} & \fcolorbox{gray}{white}{\includegraphics[width=36mm]{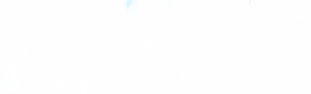}}  \\
\end{tabular}
\caption{Visual comparison of PCFA with zero-flow target on different \emph{optical flow methods} for increasing \emph{perturbation sizes} $\varepsilon_2$. White pixels represent zero flow.}
\label{fig:target_proximity}
\end{figure}

\begin{figure}[tb]
\centering
\small
\centering
\setlength{\fboxrule}{0.1pt}%
\setlength{\fboxsep}{0pt}%
\begin{tabular}{@{}r@{\ }M{36mm}@{\ \ }M{36mm}@{\ }M{36mm}@{\ }M{36mm}@{}}
            & & \footnotesize Zero-Flow & \footnotesize Negative-Flow & \footnotesize Bamboo2-41~\cite{Butler2012NaturalisticOpenSource} \\
     & \diagbox[width=36mm, height=5.6mm]{\footnotesize $\flow$ Init.}{\footnotesize Target} & \fcolorbox{gray}{white}{\includegraphics[width=36mm]{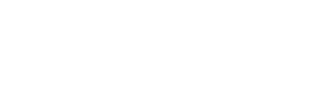}} & \fcolorbox{gray}{white}{\includegraphics[width=36mm]{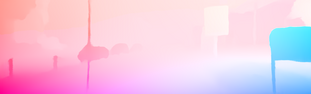}} & \fcolorbox{gray}{white}{\includegraphics[width=36mm]{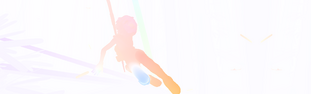}} \\
   \footnotesize FlowNet2 & \fcolorbox{gray}{white}{\includegraphics[width=36mm]{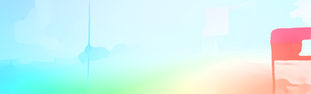}} & \fcolorbox{gray}{white}{\includegraphics[width=36mm]{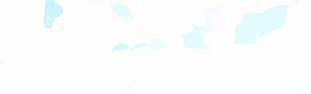}} & \fcolorbox{gray}{white}{\includegraphics[width=36mm]{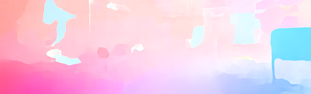}} & \fcolorbox{gray}{white}{\includegraphics[width=36mm]{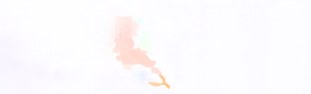}}  \\
   \footnotesize SpyNet & \fcolorbox{gray}{white}{\includegraphics[width=36mm]{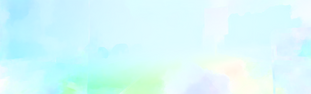}} & \fcolorbox{gray}{white}{\includegraphics[width=36mm]{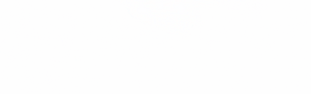}} & \fcolorbox{gray}{white}{\includegraphics[width=36mm]{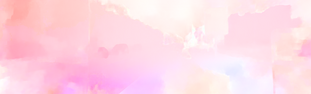}} & \fcolorbox{gray}{white}{\includegraphics[width=36mm]{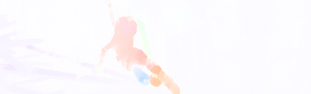}}  \\
   \footnotesize RAFT & \fcolorbox{gray}{white}{\includegraphics[width=36mm]{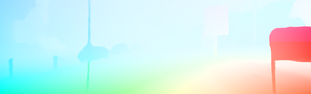}} & \fcolorbox{gray}{white}{\includegraphics[width=36mm]{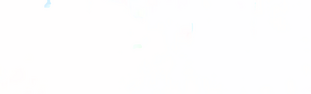}} & \fcolorbox{gray}{white}{\includegraphics[width=36mm]{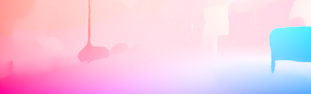}} & \fcolorbox{gray}{white}{\includegraphics[width=36mm]{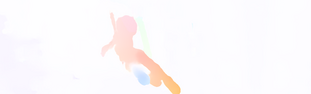}}  \\
\end{tabular}
\caption{Visual comparison of PCFA attacked flows with different \emph{targets} for multiple \emph{optical flow methods}. Choosing $\varepsilon_2 = 10^{-1}$ allows to come close to the respective target.}
\label{fig:target_selection}
\end{figure}

\medskip
\noindent\textbf{Targets.}
Next, we investigate how well PCFA can induce a given target flow for different perturbation sizes.
Fig.~\ref{fig:target_proximity_RAFT} depicts perturbed input frames, normalized adversarial perturbations and resulting flow fields for a zero-flow attack with increasing perturbation size on RAFT. Similarly, Fig.~\ref{fig:target_proximity} and Supp.\ Fig.~\ref{fig:target_proximity_app} show resulting flow fields for the other networks.
Evidently, a perturbation with larger \ltwo\ norm $\varepsilon_2$ reaches a better resemblance between the adversarial flow prediction and the all-white zero-flow target.
This is expected, as larger deviations in the input should result in larger output changes. 
However, it is remarkable that changing color values by 5\% on average ($\varepsilon_2 = 5\cdot 10^{-2}$) suffices to erase the predicted motion. 
Moreover, not only the zero-flow but also other targets can be induced with PCFA.
This is illustrated in Fig.~\ref{fig:target_selection} and Supp.\ Fig.~\ref{fig:target_selection_app}.

\medskip
\noindent\textbf{Comparison of PCFA and I-FGSM.}
Next, we compare the zero-flow performance of PCFA and I-FGSM~\cite{Schrodi2022TowardsUnderstandingAdversarial} on all networks over a range of perturbations $\varepsilon_2 \in  \{5\cdot 10^{-4},10^{-3},5\cdot 10^{-3},10^{-2}, 5\cdot 10^{-2}\}$.
We configure I-FGSM as in~\cite{Schrodi2022TowardsUnderstandingAdversarial} and use 10 optimization steps to reach $\varepsilon_\infty = \varepsilon_2$.
Note that I-FGSM does not support setting an \ltwo\ bound. 
Hence, we optimize I-FGSM for \linfty\ but track \ltwo.
\begin{figure}[tb]
\centering
   \def\svgwidth{12cm}
      \executeiffilenewer{Images/EpsPlot_predadv-target_CWFlow_FGSM_flow_zero_clipping_Kitti15.svg}{Images/EpsPlot_predadv-target_CWFlow_FGSM_flow_zero_clipping_Kitti15.pdf}%
      {inkscape -z -D --file=Images/EpsPlot_predadv-target_CWFlow_FGSM_flow_zero_clipping_Kitti15.svg --export-pdf=Images/EpsPlot_predadv-target_CWFlow_FGSM_flow_zero_clipping_Kitti15.pdf --export-latex}%
      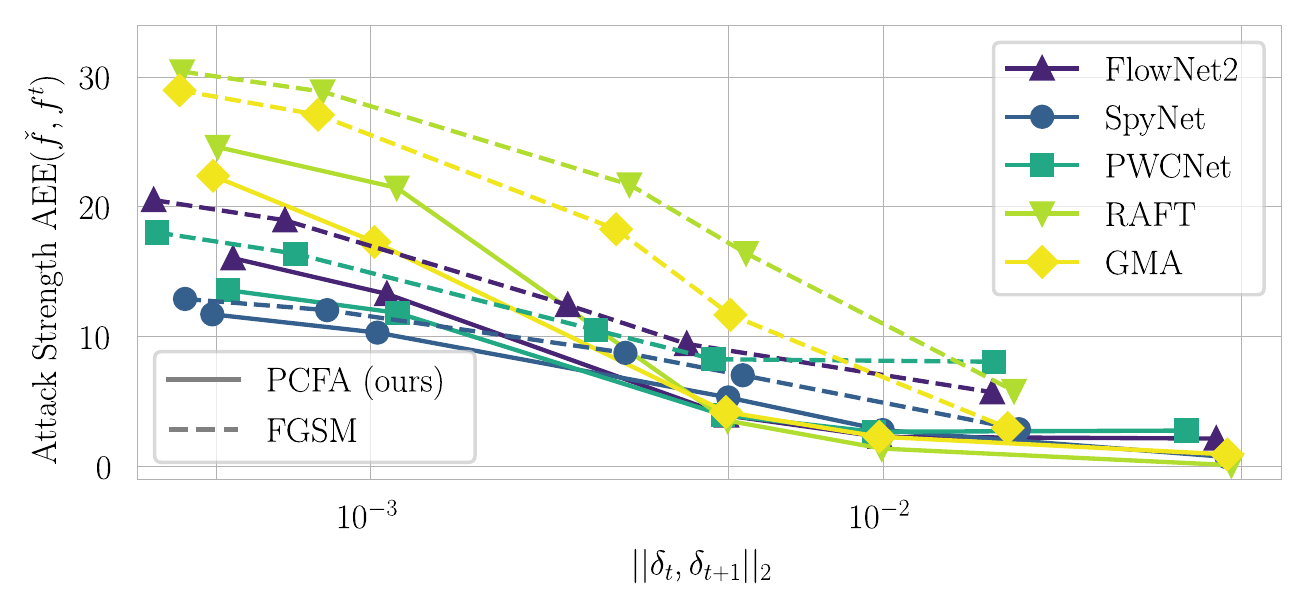%

\caption{\emph{Attack strength}
with zero-flow target over \emph{perturbation size}, for \emph{PCFA} (solid) and \emph{I-FGSM} \cite{Schrodi2022TowardsUnderstandingAdversarial} (dashed) on different flow networks. Smaller is stronger.}
\label{fig:rob_deltadelta_pt}
\end{figure}
Fig.~\ref{fig:rob_deltadelta_pt} illustrates the attack strength depending on the average perturbation norm for a zero-flow attack.
The corresponding adversarial robustness is shown in Supp.\ Fig.~\ref{fig:rob_deltadelta_pp}.
Note that the perturbation sizes for I-FGSM (dashed line) do not align well with the sampled \ltwo\ bounds for PCFA (solid line) due to its \linfty\ steering.
As one can see, PCFA achieves smaller distances between adversarial flow and target than I-FGSM -- independent of the optical
flow network and independent of the perturbation size.
Therefore, PCFA is the global attack of choice to generate strong disjoint image-specific perturbations for optical flow networks.

\subsection{Joint and Universal Perturbations}

Next we investigate PCFA's potential to generate more general, i.e.\ joint and universal, perturbations.
Moreover, to assess the transferability of network-specific joint universal perturbations, we apply them to all tested networks.

\medskip
\noindent\textbf{Joint and Universal Perturbations (White Box).}
In the white box setting, the perturbations are trained and tested on the same model and data.
We evaluate the attack strength of perturbation types as target proximity AEE$(\fadv, \targ)$ for a zero-flow attack with $\varepsilon_2 = 5\cdot 10^{-3}$ trained on the KITTI test dataset. 
For comparability, clipping is used as COV only works for disjoint perturbations.
\setlength{\tabcolsep}{4pt}
\begin{table}[tb]
\begin{center}
\caption{Zero-target proximity for different \emph{perturbations}, examples in Supp.\ Fig.~\ref{fig:deltatypes_illustr}.}
\label{table:deltatype_kitti}
\scalebox{1}{
\begin{tabular}{l@{\quad\ }l@{\quad\ }r@{\quad\ }r@{\quad\ }r@{\quad\ }r@{\quad\ }r}
\toprule
\multicolumn{2}{l}{Perturbation Type}  & FlowNet2 &  SypNet &  PWCNet &  RAFT &  GMA \\
\midrule
\multirow{2}{*}{Frame-Specific} &  $\dd$ &                \textbf{3.22} &              \textbf{4.54} &             4.28 &         \textbf{3.76} &       \textbf{3.59} \\
                           &  $\dc$ &                 4.16 &              5.84 &             \textbf{3.82} &         5.35 &       4.78 \\
\midrule
\multirow{2}{*}{Universal} & $\ddu$ &                22.03 &             14.57 &            19.13 &        28.88 &      28.49 \\
                           & $\dcu$ &                \textbf{20.49} &             \textbf{14.19} &            \textbf{18.99} &        \textbf{28.53} &      \textbf{27.17} \\
\bottomrule
\end{tabular}%
}%
\end{center}
\end{table}
\setlength{\tabcolsep}{1.4pt}

In Tab.~\ref{table:deltatype_kitti}, frame-specific perturbations clearly show a better target resemblance than universal ones for all networks, as they are optimized for frame-specific destructivity rather than transferability.
Further, disjoint perturbations are more effective than joint ones when they are frame-specific.
However, for universal perturbations the situation is reversed.
This is surprising as disjoint perturbations can adapt to both inputs, which should allow an even better target match.
While joint perturbations might add static structures to simulate zero-flow, we reproduced this observation also for a negative-flow target (\cf Supp.\ Tab.~\ref{table:deltatype_kitti_negflow}).

\medskip
\noindent\textbf{Transferability of Adversarial Perturbations (Black Box).}
To conclude the PCFA assessment, we train universal joint perturbations in a black box manner.
Per network, we optimize $\dcu$ for 25 epochs (batch size 4) on the \emph{training} set, before they are applied to the \emph{test} set for every network.
\setlength{\tabcolsep}{6pt}
\begin{table}[tb]
\begin{center}
\caption{Transferability of KITTI universal perturbations between \emph{training} and \emph{test} dataset and between different \emph{networks}, measured as adversarial robustness $\text{AEE}(\fadv,\flow)$. Large values denote a better transferability, smaller values indicate higher robustness.}
\label{table:transfer_kitti}
\begin{tabular}{l|@{\ \ \ }ccccc}
\toprule
\diagbox[width=7em, height=2em]{Test}{Train}  & FlowNet2 & SpyNet & PWCNet & RAFT & GMA \\
\midrule
FlowNet2~\cite{Ilg2017Flownet2Evolution} & \cellcolor{gray!25}\textbf{3.29} &                    \textbf{2.69} &                     2.22 &                   1.17 &                    1.12 \\
  SpyNet~\cite{Ranjan2017OpticalFlowEstimation} &                    \underline{0.60} & \cellcolor{gray!25}2.25 &                     \underline{0.57} &                   \underline{0.46} &                    \underline{0.42} \\
  PWCNet~\cite{Sun2018PwcNetCnns} &                    1.53 &                    2.19 & \cellcolor{gray!25} \textbf{2.99} &                   0.85 &                    0.75 \\
    RAFT~\cite{teed2020raft} &                    2.88 &                    \underline{1.87} &                     2.52 &\cellcolor{gray!25}3.52 &                    3.19 \\
     GMA~\cite{Jiang2021LearningEstimateHidden} &                    3.12 &                    2.14 &                     2.97 &                   \textbf{3.95} & \cellcolor{gray!25}\textbf{3.81} \\
\bottomrule
\end{tabular}
\end{center}
\end{table}
\setlength{\tabcolsep}{1.4pt}

\begin{figure}[tb]
\centering
\begin{tabular}{@{}M{12mm}@{\ }M{50mm}@{\ }M{50mm}@{\ }M{50mm}@{}}
\multicolumn{1}{l}{\textbf{KITTI}} & \includegraphics[width=50mm]{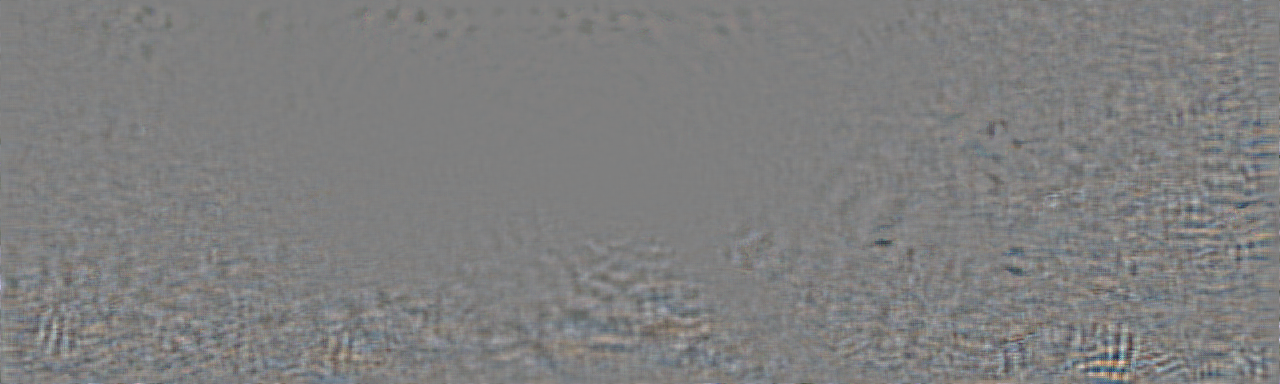} & \includegraphics[width=50mm]{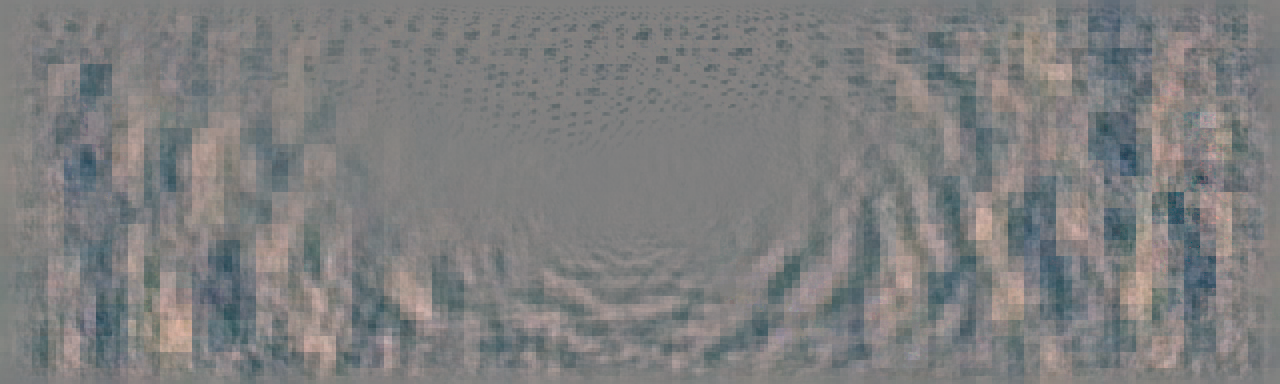} & \includegraphics[width=50mm]{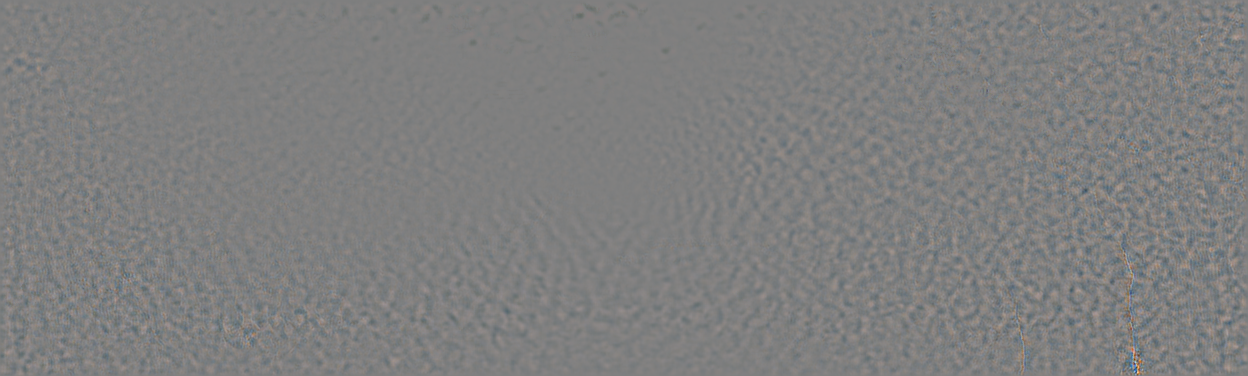}  \\
\multicolumn{1}{l}{\textbf{Sintel}} & \includegraphics[width=50mm]{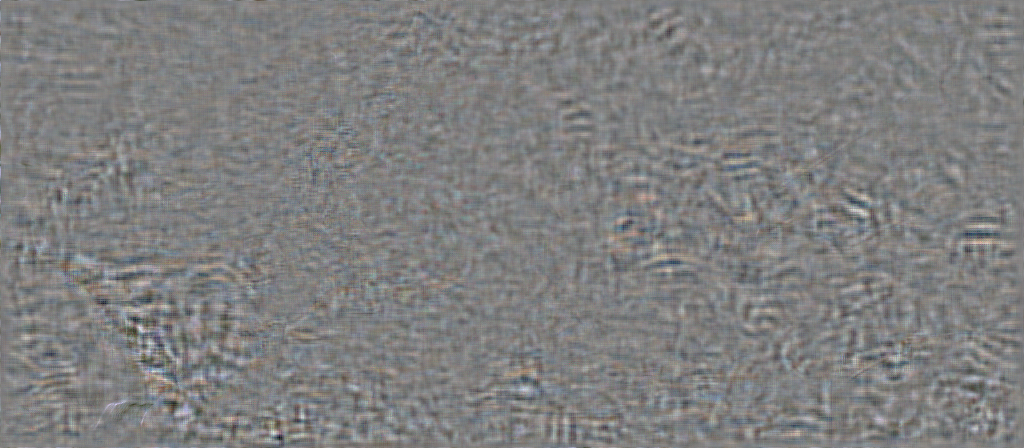} & \includegraphics[width=50mm]{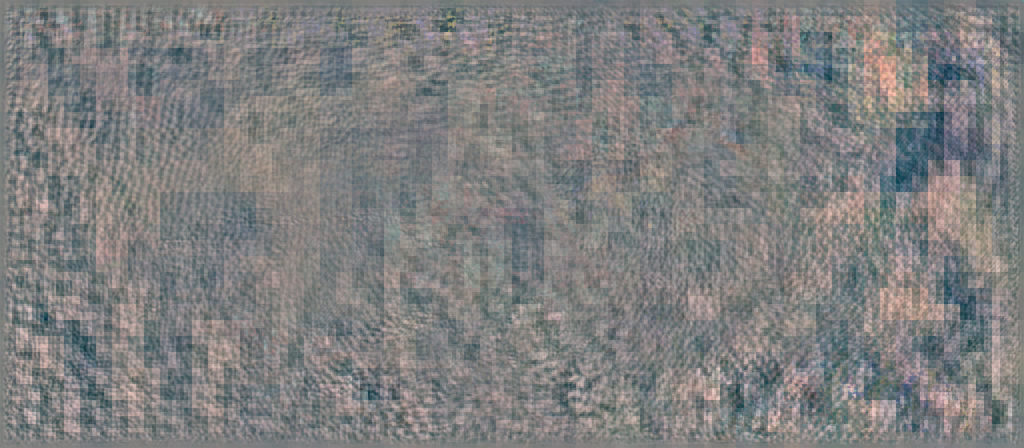} & \includegraphics[width=50mm]{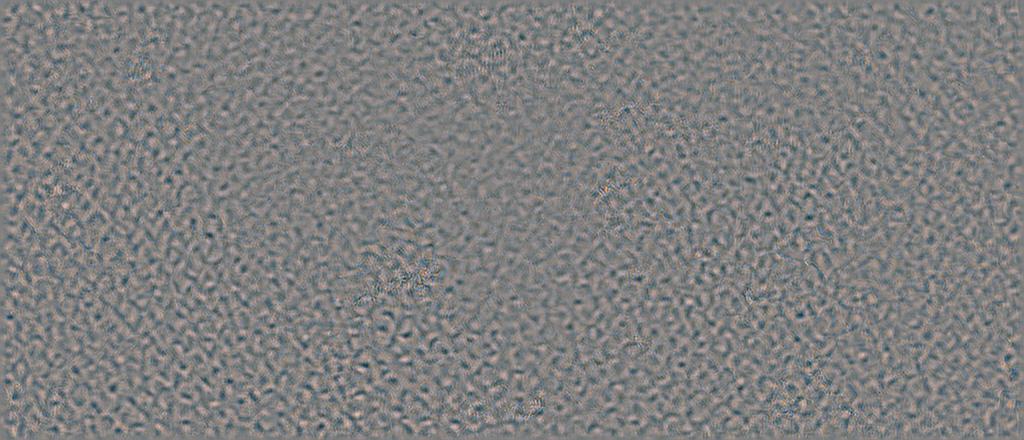}  \\
& FlowNet2~\cite{Ranjan2017OpticalFlowEstimation} & SpyNet~\cite{Ilg2017Flownet2Evolution} & RAFT~\cite{teed2020raft} \\
\end{tabular}
\caption{Normalized \emph{universal perturbations} for different network architectures learned from the respective training datasets. Top row: KITTI. Bottom row: Sintel.}
\label{fig:universal_perturb}
\end{figure}

Tab.~\ref{table:transfer_kitti} shows the adversarial robustness w.r.t.\ universal perturbations for KITTI (see Supp.\ Tab.~\ref{table:transfer_sintel} for Sintel).
Here, we observe great differences between the transferability.
While SpyNet's perturbation reliably disturbs the predictions for all networks, perturbations for RAFT or GMA mutually cause strong deviations but hardly affect other networks.
Fig.~\ref{fig:universal_perturb} and Supp.\ Fig~\ref{fig:universal_perturb_supp} further suggest that networks with transferable perturbations mainly consider well generalizing, robust features. In contrast, fine-scaled, non-transferable patterns only affect today's top methods RAFT and GMA. 
\setlength{\tabcolsep}{6pt}
\begin{table}[tb]
\begin{center}
\caption{Adversarial robustness with universal perturbations from \emph{Patch Attack}~\cite{Ranjan2019AttackingOpticalFlow} and \emph{PCFA}, with the setup from Tab.~\ref{table:transfer_kitti}. Perturbations from the KITTI training set are applied to the generating network on the KITTI test set, \cf Fig.~\ref{fig:universal_perturb} for perturbations.}
\label{table:transfer_patchPCFA}
\begin{tabular}{l@{\ \ \ }ccccc}
\toprule
Attack  & FlowNet2 & SpyNet & PWCNet & RAFT & GMA \\
\midrule
Patch Attack~\cite{Ranjan2019AttackingOpticalFlow} &   0.99 &    1.38 &    1.37 &  0.76 & 0.95 \\
PCFA (ours) & \textbf{3.29} &     \textbf{2.25} &     \textbf{2.99} &     \textbf{3.52} &      \textbf{3.81} \\
\bottomrule
\end{tabular}
\end{center}
\end{table}
\setlength{\tabcolsep}{1.4pt}%
Finally, we compare the effectiveness of the global PCFA to the Patch Attack by Ranjan \etal~\cite{Ranjan2019AttackingOpticalFlow} with \O 102 px in Tab.~\ref{table:transfer_patchPCFA}.
As both attack setups perturb a similar amount of information per frame (see supplementary material for details), this supports the initial conjecture that fewer constraints help to increase PCFA's effectiveness.

\subsection{Evaluating Quality and Robustness for Optical Flow Methods}
Finally, we use PCFA to perform a joint evaluation of adversarial robustness and optical flow quality. 
To this end, we apply PCFA's strongest configuration from Sec. 5.1, which trains $\dd$ with AAE loss and COV box constraint.
For quality, we take the official scores from KITTI and Sintel.
For robustness, we report the deviation from the initial flow for a zero-flow attack with $\varepsilon_2=5\cdot 10^{-3}$ on the respective test datasets.
 \begin{figure}[tb]
\centering
\small
\begin{tabular}[t]{@{}M{7cm}@{\ }M{7cm}@{\ }M{1.8cm}@{}}
\def\svgwidth{7cm}
      \executeiffilenewer{Images/RobQual_pred-predadv_CWFlow_0.005_zero_change_of_variables_Kitti15_bm.svg}{Images/RobQual_pred-predadv_CWFlow_0.005_zero_change_of_variables_Kitti15_bm.pdf}%
      {inkscape -z -D --file=Images/RobQual_pred-predadv_CWFlow_0.005_zero_change_of_variables_Kitti15_bm.svg --export-pdf=Images/RobQual_pred-predadv_CWFlow_0.005_zero_change_of_variables_Kitti15_bm.pdf --export-latex}%
      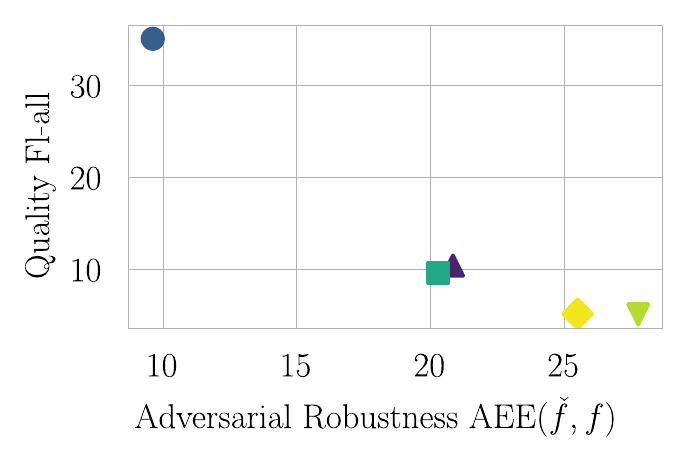%
 &
\def\svgwidth{7cm}
      \executeiffilenewer{Images/RobQual_pred-predadv_CWFlow_0.005_zero_change_of_variables_Sintel_bm.svg}{Images/RobQual_pred-predadv_CWFlow_0.005_zero_change_of_variables_Sintel_bm.pdf}%
      {inkscape -z -D --file=Images/RobQual_pred-predadv_CWFlow_0.005_zero_change_of_variables_Sintel_bm.svg --export-pdf=Images/RobQual_pred-predadv_CWFlow_0.005_zero_change_of_variables_Sintel_bm.pdf --export-latex}%
      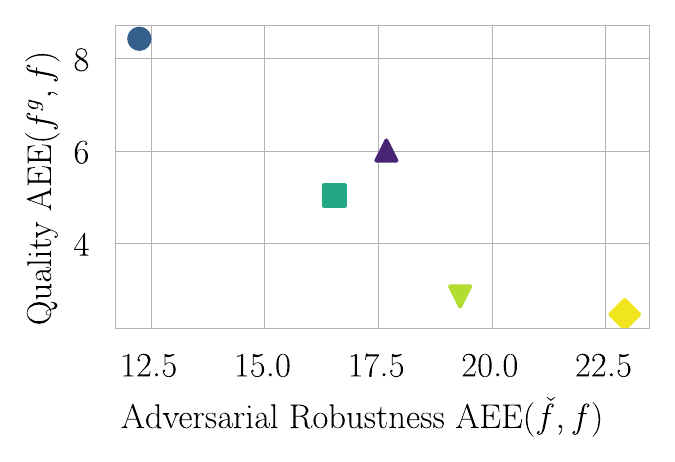%
 &
\includegraphics[width=2cm]{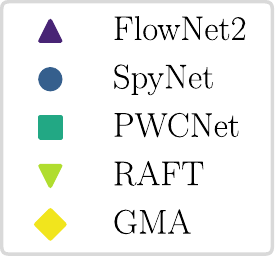} \vfill \\
\end{tabular}
\caption{Joint evaluation of optical flow methods by \emph{prediction quality} and \emph{adversarial robustness} on the respective test datasets. Left: KITTI. Right: Sintel.}
\label{fig:qual_rob_Kitti}
\end{figure}
Fig.~\ref{fig:qual_rob_Kitti} visualizes quality and adversarial robustness on different axes.
On both datasets, we observe methods with good robustness (low adversarial robustness scores) to rank bad in terms of quality (high error) and vice versa.
Further, we can identify groups of methods with similar robustness and quality scores: 
State-of-the-art networks like RAFT and GMA (recurrent) have good quality but little robustness, FlowNet2 (encoder-decoder) and PWCNet (feature pyramid) balance both quantities, while SpyNet (image pyramid) leads in robustness but offers the worst quality.
These results indicate that flow networks are subject to a trade-off between accuracy and robustness~\cite{Tsipras2019RobustnessMayBe}, which also sheds new light on the development of high accuracy methods that currently cannot sustain their top rank w.r.t.\ robustness.

%%%%%%%%%%%%%%%%%%%%%%%%%%%%%%%%%%%%%%%%%%%%%%%%%%%%%%%%%%%%%%%%%%%%%%%%%%%%%%%%%%%%%%%%%%%%%%%%%%%%%%%%%%%%
%%%%%%%%%%%%%%%%%%%%%%%%%%%%%%%%%%%%%%%%%%%%%%%%%%%%%%%%%%%%%%%%%%%%%%%%%%%%%%%%%%%%%%%%%%%%%%%%%%%%%%%%%%%%
%%%%
%%%%  DISCUSSION / CONCLUSION
%%%%
%%%%%%%%%%%%%%%%%%%%%%%%%%%%%%%%%%%%%%%%%%%%%%%%%%%%%%%%%%%%%%%%%%%%%%%%%%%%%%%%%%%%%%%%%%%%%%%%%%%%%%%%%%%%
%%%%%%%%%%%%%%%%%%%%%%%%%%%%%%%%%%%%%%%%%%%%%%%%%%%%%%%%%%%%%%%%%%%%%%%%%%%%%%%%%%%%%%%%%%%%%%%%%%%%%%%%%%%%

\section{Conclusions}

This work describes the Perturbation-Constrained Flow Attack (PCFA), a novel global adversarial attack designed for a rigorous adversarial robustness assessment of optical flow networks.
In contrast to previous flow attacks, PCFA finds more destructive adversarial perturbations and effectively limits their \ltwo\ norm, which renders it particularly suitable for comparing the robustness of neural networks.
Our experimental analysis clearly shows that high quality flow methods are not automatically robust. In fact, these methods seem to be particularly vulnerable to PCFA's perturbations.
Therefore, we strongly encourage the research community to treat robustness with equal importance as quality and report both metrics for optical flow methods.
With PCFA we not only provide a systematic tool to do so, but with our formal definition of adversarial robustness we also provide a general concept that allows to compare both methods and attacks.

\section*{Acknowledgments}
This work was funded by the Deutsche Forschungsgemeinschaft (DFG, German Research Foundation) -- Project-ID 251654672 -- TRR 161 (B04). Jenny Schmalfuss is supported by the International Max Planck Research School for Intelligent Systems (IMPRS-IS).

\bibliographystyle{unsrt}
\bibliography{bibliography}

\cleardoublepage

%%%%%%%%%%%%%%%%%%%%%%%%%%%%%%%%%%%%%%%%
%
%  Supplementary Material
%
%%%%%%%%%%%%%%%%%%%%%%%%%%%%%%%%%%%%%%%%

% Starts a separate bibliography counter for the supplementary material
\begin{bibunit}[unsrt]

\title{Supplementary Material}
\author{}
\settitle

\vspace{-7\baselineskip}

\appendix
\setcounter{page}{1}
\setcounter{table}{0}
\renewcommand{\thetable}{A\arabic{table}}
\setcounter{figure}{0}
\renewcommand{\thefigure}{A\arabic{figure}} 

\section{Experiment Configurations}

Tab.~\ref{table:exp_configs} summarizes the experimental configurations, where different visualizations or representations of the same experiment class are grouped together.
If several cells contain multiple choices, the experiment was conducted for all possible combinations of those.
The only exception are $\varepsilon_2$ and $\mu$, which should match by line.

The attacks are evaluated on the networks FlowNet2~\cite{Ilg2017Flownet2Evolution} (implementation from \cite{Reda2017Flownet2PytorchPytorch}), PWCNet~\cite{Sun2018PwcNetCnns}, SpyNet~\cite{Ranjan2017OpticalFlowEstimation} (implementation from~\cite{Niklaus2018PytorchSpyNet}), RAFT~\cite{teed2020raft} and GMA~\cite{Jiang2021LearningEstimateHidden}.
All network implementations use checkpoints that are not fine-tuned to the KITTI15~\cite{Menze2015Joint3dEstimation} data, other checkpoints and networks can easily be evaluated with our implementation (\href{https://github.com/cv-stuttgart/PCFA}{https://github.com/cv-stuttgart/PCFA}).

\setlength{\tabcolsep}{4pt}
\begin{table}[h]
\begin{center}
\caption{Parameters and Configurations for the experimental Results in Section~\ref{sec:Experiments}. The evaluation- and test splits of the KITTI15 dataset~\cite{Menze2015Joint3dEstimation} are denoted \emph{K15-te} and \emph{K15-tr} respectively; For the MPI-Sintel dataset~\cite{Butler2012NaturalisticOpenSource} we use \emph{S-te} and \emph{S-tr}.}
\label{table:exp_configs}
\scalebox{0.8}{
\begin{tabular}{M{12mm}M{12mm}M{20mm}M{5mm}M{12mm}M{12mm}M{10mm}M{12mm}M{10mm}M{10mm}M{10mm}M{10mm}M{10mm}M{10mm}M{10mm}M{10mm}M{10mm}M{10mm}M{10mm}M{10mm}M{10mm}M{10mm}M{10mm}M{10mm}}
\toprule
\multicolumn{1}{l}{Experiment} & Attack & Network & $\targ$ & Loss & Box Constr. & Perturb. Type  & $\varepsilon_2$ & Penalty $\mu$ & Optim. Steps & Batch Size & Epo. & Dataset (train) & Dataset (eval)\\
\midrule
\multicolumn{1}{l}{Tab.~\ref{table:setup_PCFA_loss_tgt_pt},\ref{table:setup_PCFA_loss_tgt_pp}} & PCFA & RAFT & $0$, $-\flow$ & AEE      & Clip, COV & $\dd$ &  $5 \cdot 10^{-3}$ & $5\cdot 10^5$ & 20 & 1 & 1 & K15-te & K15-te  \\ 
\multicolumn{1}{l}{Tab.~\ref{table:setup_PCFA_loss_tgt_pt},\ref{table:setup_PCFA_loss_tgt_pp}} & PCFA & RAFT & $0$           & MSE, CS  & Clip, COV & $\dd$ &  $5 \cdot 10^{-3}$ & $5\cdot 10^6$ & 20 & 1 & 1 & K15-te & K15-te \\ 
\multicolumn{1}{l}{Tab.~\ref{table:setup_PCFA_loss_tgt_pt},\ref{table:setup_PCFA_loss_tgt_pp}} & PCFA & RAFT &      $-\flow$ & MSE, CS  & Clip, COV & $\dd$ &  $5 \cdot 10^{-3}$ & $7\cdot 10^6$ & 20 & 1 & 1 & K15-te & K15-te \\ 
\midrule
\multicolumn{1}{l}{Fig.~\ref{fig:target_proximity_RAFT},\ref{fig:target_proximity},\ref{fig:target_proximity_app}} & PCFA & FlowNet2, PWCNet, SpyNet, GMA, RAFT & $0$  & AEE      & COV & $\dd$ & $5 \cdot 10^{-2}$ $5 \cdot 10^{-3}$ $5 \cdot 10^{-4}$ & $5\cdot 10^4$ $5\cdot 10^5$ $5\cdot 10^6$ & 20 & 1 & 1 & K15-te & K15-te \\ 
\midrule
\multicolumn{1}{l}{Fig.~\ref{fig:target_selection},\ref{fig:target_selection_app}} & PCFA & FlowNet2, PWCNet, SpyNet, GMA, RAFT & $0$ $-\flow$ B2-41  & AEE      & COV & $\dd$ & $1 \cdot 10^{-1}$ $1 \cdot 10^{-1}$ $1 \cdot 10^{-1}$ & $1\cdot 10^4$ $2\cdot 10^4$ $2\cdot 10^4$ & 20 & 1 & 1 & K15-te & K15-te \\ 
\midrule
\multicolumn{1}{l}{Fig.~\ref{fig:rob_deltadelta_pt},~\ref{fig:rob_deltadelta_pp}} & PCFA & FlowNet2, PWCNet, SpyNet, GMA, RAFT & $0$  & AEE      & COV & $\dd$ & $5 \cdot 10^{-2}$ $1 \cdot 10^{-2}$ $5 \cdot 10^{-3}$ $1 \cdot 10^{-3}$ $5 \cdot 10^{-4}$ & $5\cdot 10^4$ $1\cdot 10^5$ $5\cdot 10^5$ $1\cdot 10^6$ $5\cdot 10^6$ & 20 & 1 & 1 & K15-te & K15-te \\ 
\multicolumn{1}{l}{Fig.~\ref{fig:rob_deltadelta_pt},~\ref{fig:rob_deltadelta_pp}} & I-FGSM & FlowNet2, PWCNet, SpyNet, GMA, RAFT & $0$  & AEE      & COV & $\dd$ & [$\varepsilon_\infty$] $5 \cdot 10^{-2}$ $1 \cdot 10^{-2}$ $5 \cdot 10^{-3}$ $1 \cdot 10^{-3}$ $5 \cdot 10^{-4}$ & \emph{n.a.} & 10 & 1 & 1 & K15-te & K15-te \\ 
\midrule
\multicolumn{1}{l}{Tab.~\ref{table:deltatype_kitti}, Fig.~\ref{fig:deltatypes_illustr}} & PCFA & FlowNet2, PWCNet, SpyNet, GMA, RAFT & $0$  & AEE      & Clip & $\dd$ $\dc$   & $5 \cdot 10^{-3}$ & $5\cdot 10^5$ & 20 & 1 & 1 & K15-te & K15-te \\ 
\multicolumn{1}{l}{Tab.~\ref{table:deltatype_kitti}, Fig.~\ref{fig:deltatypes_illustr}} & PCFA & FlowNet2, PWCNet, SpyNet, GMA, RAFT & $0$  & AEE      & Clip & $\ddu$ $\dcu$ & $5 \cdot 10^{-3}$ & $5\cdot 10^5$ & 1 & 4 & 25 & K15-te & K15-te \\ 
\multicolumn{1}{l}{Tab.~\ref{table:deltatype_kitti_negflow}} & PCFA & FlowNet2, PWCNet, SpyNet, GMA, RAFT & $-\flow$  & AEE      & Clip & $\dd$ $\dc$ & $5 \cdot 10^{-3}$ & $1\cdot 10^6$ & 20 & 1 & 1 & K15-te & K15-te \\
\multicolumn{1}{l}{Tab.~\ref{table:deltatype_kitti_negflow}} & PCFA & FlowNet2, PWCNet, SpyNet, GMA, RAFT & $-\flow$  & AEE      & Clip & $\ddu$ $\dcu$ & $5 \cdot 10^{-3}$ & $1\cdot 10^6$ & 1 & 4 & 25 & K15-te & K15-te \\
\midrule
\multicolumn{1}{l}{Tab.~\ref{table:transfer_kitti}, Fig.~\ref{fig:universal_perturb}} & PCFA & FlowNet2, PWCNet, SpyNet, GMA, RAFT & $0$  & AEE      & Clip & $\dcu$ & $5 \cdot 10^{-3}$ & $5\cdot 10^5$ & 1 & 4 & 25 & K15-tr & K15-te \\ 
\multicolumn{1}{l}{Tab.~\ref{table:transfer_sintel}, Fig.~\ref{fig:universal_perturb_supp}} & PCFA & FlowNet2, PWCNet, SpyNet, GMA, RAFT & $0$  & AEE      & Clip & $\dcu$ & $5 \cdot 10^{-3}$ & $5\cdot 10^5$ & 1 & 4 & 25 & S-tr & S-te \\ 
\bottomrule
\end{tabular}
}
\end{center}
\end{table}
\setlength{\tabcolsep}{1.4pt}

\section{Additional Material}

\subsection{Additional Results for PCFA on Specific Frame Pairs}

To complement the configuration study for PCFA that reported the attack strength in Main Tab.~\ref{table:setup_PCFA_loss_tgt_pt}, we additionally provide the \emph{adversarial robustness} measures for the tested configurations in Tab~\ref{table:setup_PCFA_loss_tgt_pp}.
For the zero-flow target and the cosine similarity (\emph{CS}) loss, no deviation between the inital and the adversarial flow is induced, which shows that the cosine similarity does not train the perturbation towards the zero target.
For the negative-flow target, it appears that greater deviations between adversarial and initial flow can be induced by the cosine similarity with clipping.
However, a comparison to the target proximity in Main Tab.~\ref{table:setup_PCFA_loss_tgt_pt} clearly shows that the high deviation between adversarial and initial flow stems from a non-converging method, i.e.\ a very large distance to target, rather than from a strong targeted approach.
\setlength{\tabcolsep}{4pt}
\begin{table}
\begin{center}
\caption{PCFA adversarial robustness $\text{AEE}(\fadv, \flow)$ for different \emph{loss functions}, \emph{targets} and \emph{box constraints} on RAFT. Larger values indicate a bigger deviation between adversarial and initial flow.}
\label{table:setup_PCFA_loss_tgt_pp}
\begin{tabular}{@{}r@{\quad\quad}rrr@{\quad\quad}rrr@{}}
\toprule
          &  \multicolumn{3}{c}{$\targ = 0$\phantom{aaaa}}         &  \multicolumn{3}{c}{$\targ = -\flow$}                \\
          \cmidrule(l{0.1em}r{2.1em}){2-4}  \cmidrule(l{0.2em}r{0em}){5-7}
         &  AEE &  MSE &  CS &  AEE &  MSE &  CS \\
\midrule
Clipping &        29.12 &     22.96 &         0.00 &         44.11 &         29.09 &    \textbf{81.35} \\
     COV &\textbf{29.31} &     25.88 &         0.00 &         47.99 &         31.94 &            40.19 \\
\bottomrule
\end{tabular}
\end{center}
\end{table}
\setlength{\tabcolsep}{1.4pt}

\begin{figure}
\centering
\setlength{\fboxrule}{0.1pt}%
\setlength{\fboxsep}{0pt}%
\begin{tabular}{@{}r@{\ }M{36mm}@{\ \ \ }M{36mm}@{\ }M{36mm}@{\ }M{36mm}@{}}
            & Initial Flow & $\varepsilon_2=5\cdot 10^{-4}$ & $\varepsilon_2=5\cdot 10^{-3}$ & $\varepsilon_2=5\cdot 10^{-2}$ \\
   \footnotesize FlowNet2 & \fcolorbox{gray}{white}{\includegraphics[width=36mm]{target_progression/FlowNet2_zero_CWFlow_0_0_flow.png}} & \fcolorbox{gray}{white}{\includegraphics[width=36mm]{target_progression/FlowNet2_zero_CWFlow_0_0.0005_flow.png}} & \fcolorbox{gray}{white}{\includegraphics[width=36mm]{target_progression/FlowNet2_zero_CWFlow_0_0.005_flow.png}} & \fcolorbox{gray}{white}{\includegraphics[width=36mm]{target_progression/FlowNet2_zero_CWFlow_0_0.05_flow.png}}  \\
   PWCNet & \fcolorbox{gray}{white}{\includegraphics[width=36mm]{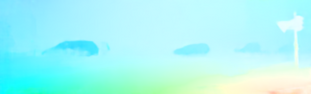}} & \fcolorbox{gray}{white}{\includegraphics[width=36mm]{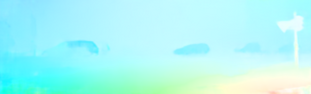}} & \fcolorbox{gray}{white}{\includegraphics[width=36mm]{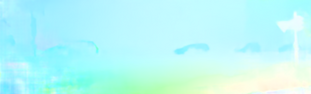}} & \fcolorbox{gray}{white}{\includegraphics[width=36mm]{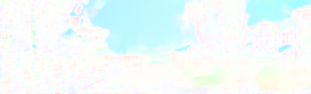}}  \\
   \footnotesize SpyNet & \fcolorbox{gray}{white}{\includegraphics[width=36mm]{target_progression/SpyNet_zero_CWFlow_0_0_flow.png}} & \fcolorbox{gray}{white}{\includegraphics[width=36mm]{target_progression/SpyNet_zero_CWFlow_0_0.0005_flow.png}} & \fcolorbox{gray}{white}{\includegraphics[width=36mm]{target_progression/SpyNet_zero_CWFlow_0_0.005_flow.png}} & \fcolorbox{gray}{white}{\includegraphics[width=36mm]{target_progression/SpyNet_zero_CWFlow_0_0.05_flow.png}}  \\
    RAFT & \fcolorbox{gray}{white}{\includegraphics[width=36mm]{target_progression/RAFT_zero_CWFlow_0_0_flow.png}} & \fcolorbox{gray}{white}{\includegraphics[width=36mm]{target_progression/RAFT_zero_CWFlow_0_0.0005_flow.png}} & \fcolorbox{gray}{white}{\includegraphics[width=36mm]{target_progression/RAFT_zero_CWFlow_0_0.005_flow.png}} & \fcolorbox{gray}{white}{\includegraphics[width=36mm]{target_progression/RAFT_zero_CWFlow_0_0.05_flow.png}}  \\
   \footnotesize GMA & \fcolorbox{gray}{white}{\includegraphics[width=36mm]{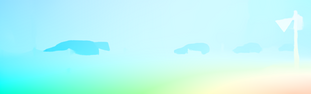}} & \fcolorbox{gray}{white}{\includegraphics[width=36mm]{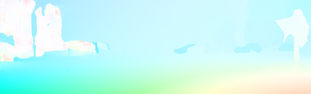}} & \fcolorbox{gray}{white}{\includegraphics[width=36mm]{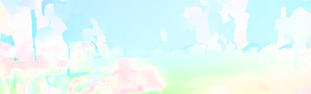}} & \fcolorbox{gray}{white}{\includegraphics[width=36mm]{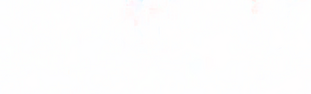}}  \\[24pt]
   FlowNet2 & \fcolorbox{gray}{white}{\includegraphics[width=36mm]{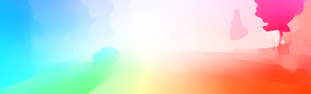}} & \fcolorbox{gray}{white}{\includegraphics[width=36mm]{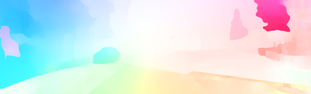}} & \fcolorbox{gray}{white}{\includegraphics[width=36mm]{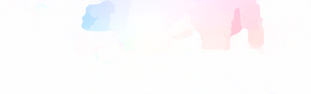}} & \fcolorbox{gray}{white}{\includegraphics[width=36mm]{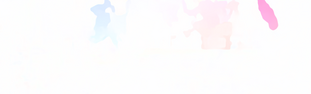}}  \\
   PWCNet & \fcolorbox{gray}{white}{\includegraphics[width=36mm]{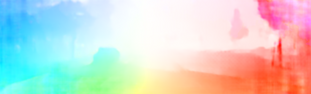}} & \fcolorbox{gray}{white}{\includegraphics[width=36mm]{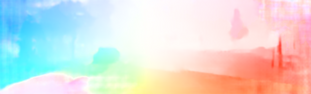}} & \fcolorbox{gray}{white}{\includegraphics[width=36mm]{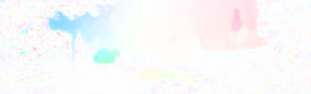}} & \fcolorbox{gray}{white}{\includegraphics[width=36mm]{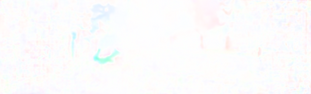}}  \\
   SpyNet & \fcolorbox{gray}{white}{\includegraphics[width=36mm]{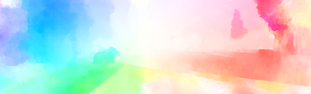}} & \fcolorbox{gray}{white}{\includegraphics[width=36mm]{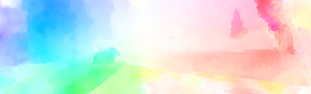}} & \fcolorbox{gray}{white}{\includegraphics[width=36mm]{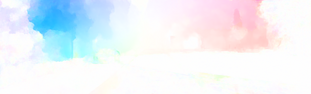}} & \fcolorbox{gray}{white}{\includegraphics[width=36mm]{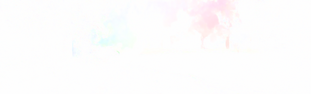}}  \\
   RAFT & \fcolorbox{gray}{white}{\includegraphics[width=36mm]{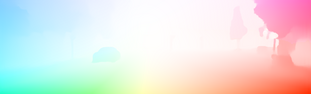}} & \fcolorbox{gray}{white}{\includegraphics[width=36mm]{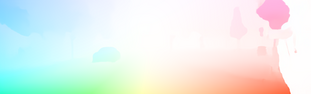}} & \fcolorbox{gray}{white}{\includegraphics[width=36mm]{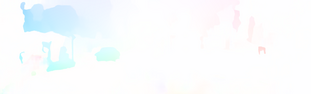}} & \fcolorbox{gray}{white}{\includegraphics[width=36mm]{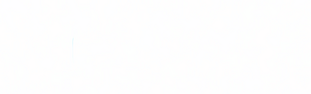}}  \\
   GMA & \fcolorbox{gray}{white}{\includegraphics[width=36mm]{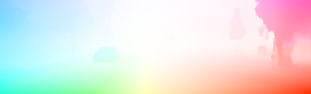}} & \fcolorbox{gray}{white}{\includegraphics[width=36mm]{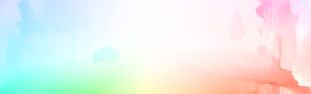}} & \fcolorbox{gray}{white}{\includegraphics[width=36mm]{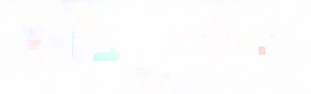}} & \fcolorbox{gray}{white}{\includegraphics[width=36mm]{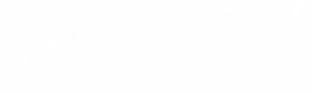}}  \\
\end{tabular}
\caption{Visual comparison of PCFA with zero-flow target on different \emph{optical flow methods} for increasing \emph{perturbation sizes} $\varepsilon_2$ on two exemplary scenes from KITTI15. White pixels represent the zero flow.}
\label{fig:target_proximity_app}
\end{figure}

\begin{figure}[tb]
\centering
\small
\centering
\setlength{\fboxrule}{0.1pt}%
\setlength{\fboxsep}{0pt}%
\begin{tabular}{@{}r@{\ }M{36mm}@{\ \ \ }M{36mm}@{\ }M{36mm}@{\ }M{36mm}@{}}
   % \hline
            & & \footnotesize Zero-Flow & \footnotesize Negative-Flow & \footnotesize Bamboo2-41 \\
     & \diagbox[width=36mm, height=5.7mm]{\footnotesize $\flow$ Init.}{\footnotesize Target} & \fcolorbox{gray}{white}{\includegraphics[width=36mm]{targets_0.1/RAFT_zero_CWFlow_2_0.1_flow_target}} & \fcolorbox{gray}{white}{\includegraphics[width=36mm]{targets_0.1/RAFT_neg_flow_CWFlow_2_0.1_flow_target.png}} & \fcolorbox{gray}{white}{\includegraphics[width=36mm]{targets_0.1/RAFT_custom_CWFlow_2_0.1_flow_target.png}} \\[12pt]
   \footnotesize FlowNet2 & \fcolorbox{gray}{white}{\includegraphics[width=36mm]{targets_0.1/FlowNet2_zero_CWFlow_2_0.1_flow_init.png}} & \fcolorbox{gray}{white}{\includegraphics[width=36mm]{targets_0.1/FlowNet2_zero_CWFlow_2_0.1_flow_adv.png}} & \fcolorbox{gray}{white}{\includegraphics[width=36mm]{targets_0.1/FlowNet2_neg_flow_CWFlow_2_0.1_flow_adv.png}} & \fcolorbox{gray}{white}{\includegraphics[width=36mm]{targets_0.1/FlowNet2_custom_CWFlow_2_0.1_flow_adv.png}}  \\
   \footnotesize PWCNet & \fcolorbox{gray}{white}{\includegraphics[width=36mm]{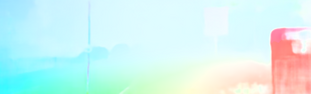}} & \fcolorbox{gray}{white}{\includegraphics[width=36mm]{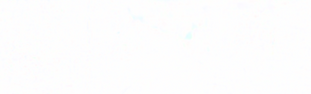}} & \fcolorbox{gray}{white}{\includegraphics[width=36mm]{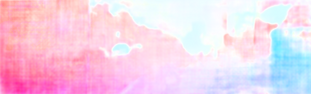}} & \fcolorbox{gray}{white}{\includegraphics[width=36mm]{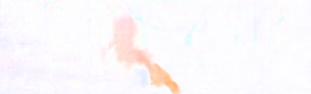}}  \\
   \footnotesize SpyNet & \fcolorbox{gray}{white}{\includegraphics[width=36mm]{targets_0.1/SpyNet_zero_CWFlow_2_0.1_flow_init.png}} & \fcolorbox{gray}{white}{\includegraphics[width=36mm]{targets_0.1/SpyNet_zero_CWFlow_2_0.1_flow_adv.png}} & \fcolorbox{gray}{white}{\includegraphics[width=36mm]{targets_0.1/SpyNet_neg_flow_CWFlow_2_0.1_flow_adv.png}} & \fcolorbox{gray}{white}{\includegraphics[width=36mm]{targets_0.1/SpyNet_custom_CWFlow_2_0.1_flow_adv.png}}  \\
   \footnotesize RAFT & \fcolorbox{gray}{white}{\includegraphics[width=36mm]{targets_0.1/RAFT_zero_CWFlow_2_0.1_flow_init.png}} & \fcolorbox{gray}{white}{\includegraphics[width=36mm]{targets_0.1/RAFT_zero_CWFlow_2_0.1_flow_adv.png}} & \fcolorbox{gray}{white}{\includegraphics[width=36mm]{targets_0.1/RAFT_neg_flow_CWFlow_2_0.1_flow_adv.png}} & \fcolorbox{gray}{white}{\includegraphics[width=36mm]{targets_0.1/RAFT_custom_CWFlow_2_0.1_flow_adv.png}}  \\
   \footnotesize GMA & \fcolorbox{gray}{white}{\includegraphics[width=36mm]{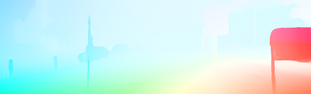}} & \fcolorbox{gray}{white}{\includegraphics[width=36mm]{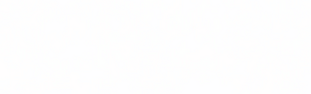}} & \fcolorbox{gray}{white}{\includegraphics[width=36mm]{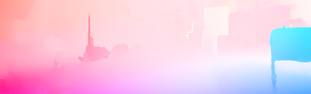}} & \fcolorbox{gray}{white}{\includegraphics[width=36mm]{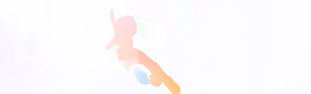}}  \\[24pt]
            & & \footnotesize Zero-Flow & \footnotesize Negative-Flow & \footnotesize Bamboo2-41 \\
   & \diagbox[width=36mm, height=5.7mm]{\footnotesize $\flow$ Init.}{\footnotesize Target} & \fcolorbox{gray}{white}{\includegraphics[width=36mm]{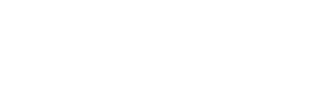}} & \fcolorbox{gray}{white}{\includegraphics[width=36mm]{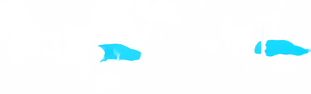}} & \fcolorbox{gray}{white}{\includegraphics[width=36mm]{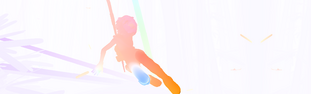}} \\[12pt]
   \footnotesize FlowNet2 & \fcolorbox{gray}{white}{\includegraphics[width=36mm]{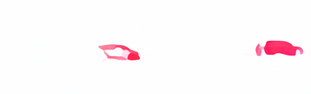}} & \fcolorbox{gray}{white}{\includegraphics[width=36mm]{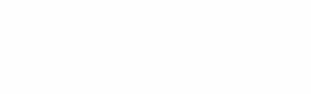}} & \fcolorbox{gray}{white}{\includegraphics[width=36mm]{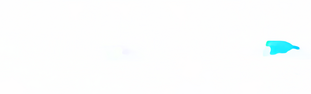}} & \fcolorbox{gray}{white}{\includegraphics[width=36mm]{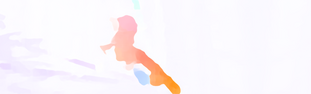}}  \\
   \footnotesize PWCNet & \fcolorbox{gray}{white}{\includegraphics[width=36mm]{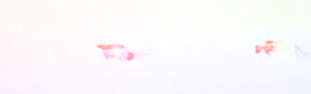}} & \fcolorbox{gray}{white}{\includegraphics[width=36mm]{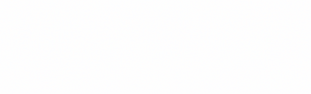}} & \fcolorbox{gray}{white}{\includegraphics[width=36mm]{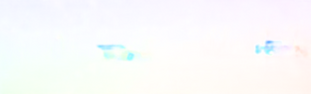}} & \fcolorbox{gray}{white}{\includegraphics[width=36mm]{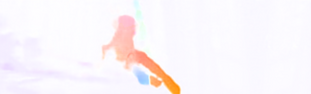}}  \\
   \footnotesize SpyNet & \fcolorbox{gray}{white}{\includegraphics[width=36mm]{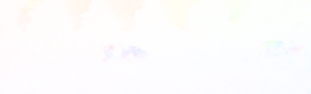}} & \fcolorbox{gray}{white}{\includegraphics[width=36mm]{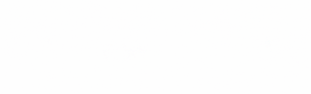}} & \fcolorbox{gray}{white}{\includegraphics[width=36mm]{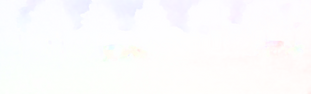}} & \fcolorbox{gray}{white}{\includegraphics[width=36mm]{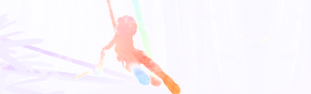}}  \\
   \footnotesize RAFT & \fcolorbox{gray}{white}{\includegraphics[width=36mm]{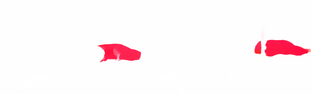}} & \fcolorbox{gray}{white}{\includegraphics[width=36mm]{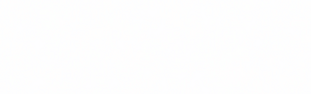}} & \fcolorbox{gray}{white}{\includegraphics[width=36mm]{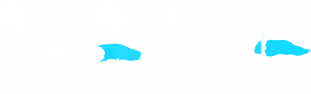}} & \fcolorbox{gray}{white}{\includegraphics[width=36mm]{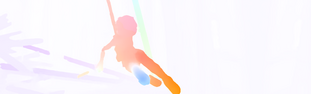}}  \\
   \footnotesize GMA & \fcolorbox{gray}{white}{\includegraphics[width=36mm]{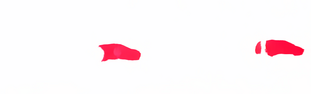}} & \fcolorbox{gray}{white}{\includegraphics[width=36mm]{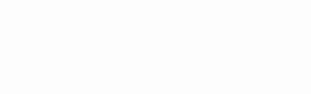}} & \fcolorbox{gray}{white}{\includegraphics[width=36mm]{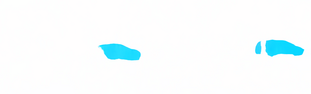}} & \fcolorbox{gray}{white}{\includegraphics[width=36mm]{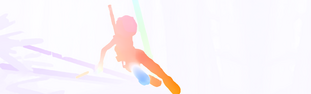}}  \\
   % \hline
\end{tabular}
\caption{Visual comparison of the result of PCFA with different \emph{targets} on different \emph{optical flow methods}. Choosing $\varepsilon_2 = 10^{-1}$ allows to come close to any target. \emph{Bamboo2-41} is a ground truth flow from Sintel final's Bamboo2 sequence.}
\label{fig:target_selection_app}
\end{figure}

In the main paper we visualize the effect of increasingly large perturbations on the attacked flow in Main Fig.~\ref{fig:target_proximity_RAFT} and Main Fig.~\ref{fig:target_proximity} for the networks FlowNet2, SpyNet and RAFT.
The results for all networks and for an additional input frame pair are shown in Fig.~\ref{fig:target_proximity_app}.
Similarly, Fig.~\ref{fig:target_selection_app} complements the reduced selection from Main Fig.~\ref{fig:target_selection} with additional illustrations of adversarial flows for the chosen targets on all tested networks.

\begin{figure}[tb]
\centering,
   \def\svgwidth{12cm}
      \executeiffilenewer{Images/EpsPlot_pred-predadv_CWFlow_FGSM_flow_zero_clipping_Kitti15.svg}{Images/EpsPlot_pred-predadv_CWFlow_FGSM_flow_zero_clipping_Kitti15.pdf}%
      {inkscape -z -D --file=Images/EpsPlot_pred-predadv_CWFlow_FGSM_flow_zero_clipping_Kitti15.svg --export-pdf=Images/EpsPlot_pred-predadv_CWFlow_FGSM_flow_zero_clipping_Kitti15.pdf --export-latex}%
      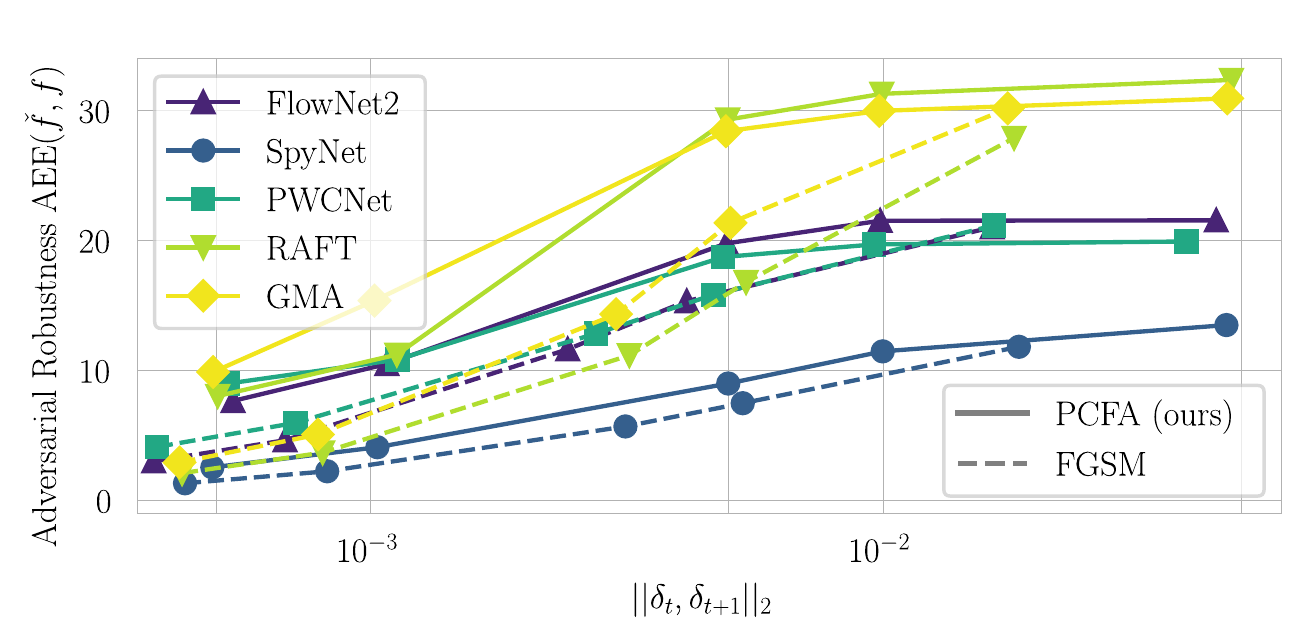%

\caption{\emph{Adversarial robustness} with zero-flow target over \emph{perturbation size}, for \emph{PCFA} (solid) and \emph{I-FGSM}~\cite{Schrodi2022TowardsUnderstandingAdversarial,Kurakin2017AdversarialMachineLearning} (dashed) on different flow networks. Larger indicates a greater distance of the adversarial from the initial flow.}
\label{fig:rob_deltadelta_pp}
\end{figure}
In addition to Main Fig.~\ref{fig:rob_deltadelta_pt}, Fig.~\ref{fig:rob_deltadelta_pp} visualizes the distance between adversarial and initial flow that is induced by perturbations with increasing size generated with PCFA and I-FGSM~\cite{Schrodi2022TowardsUnderstandingAdversarial,Kurakin2017AdversarialMachineLearning}.
For large perturbation sizes ($\varepsilon_2 \geq 10^{-2}$), it appears that I-FGSM can cause the adversarial robustness (Fig.~\ref{fig:rob_deltadelta_pp}) to degrade to a similar extent than PCFA. However, comparing to the target proximity in Main Fig.~\ref{fig:rob_deltadelta_pt} again shows that the target proximity that is reached for I-FGSM is not as good as the one reached by PCFA.
Consequently, I-FGSM perturbs the flow \emph{away} from the initial flow, but does so in an untargeted manner as it fails to resemble the target flow.
Meanwhile, PCFA has mostly converged to the zero flow for large perturbations and hence does not induce a further strong change, which explains why the distance between adversarial and initial flow does not increase further.
Therefore, these results support that PCFA is a stronger method, and better suited than I-FGSM to generate perturbations that induce a desired target flow.

\subsection{Additional Results for Joint and Universal Perturbations}

\subsubsection{Joint and Universal Perturbations.} Tab.~\ref{table:deltatype_kitti_negflow} extends the results in Main Tab.~\ref{table:deltatype_kitti} by approximating the negative-flow target instead of the zero-flow.
Both tables show the corresponding target proximity for different perturbation types generated with PCFA.
Because the trend in Tab.~\ref{table:deltatype_kitti_negflow} agrees with Main Tab.~\ref{table:deltatype_kitti}, where joint universal perturbations reach a better target resemblance than disjoint ones, both results suggests that the greater effectiveness of joint over disjoint universal perturbations is not related to the used target.
A possible explanation for the better performance of the universal joint perturbations is that the batched training of universal perturbations tends to overfit on the batches, and may therefore not be able to learn better generalizing perturbations.
In contrast, training one joint perturbation for both images explicitly incorporates a type of generalization (over input frames) into the optimization.
\setlength{\tabcolsep}{4pt}
\begin{table}
\begin{center}
\caption{Negative-flow target proximity for different \emph{perturbations}.}
\label{table:deltatype_kitti_negflow}
\scalebox{1}{
\begin{tabular}{l@{\quad\ }l@{\quad\ }r@{\quad\ }r@{\quad\ }r@{\quad\ }r@{\quad\ }r}
\toprule
\multicolumn{2}{l}{Perturbation Type}  & FlowNet2 &  SypNet &  PWCNet &  RAFT &  GMA \\
\midrule
\multirow{2}{*}{Frame-Specific} & $\dd$  & \textbf{14.73} & \textbf{15.33} & \textbf{13.36} & \textbf{19.86} & \textbf{16.74} \\
                                & $\dc$  & 19.07 & 18.54 & 15.60 & 30.35 & 20.48 \\
\midrule
\multirow{2}{*}{Universal} & $\ddu$  & 44.97 & 29.93 & 40.50 & 61.04 & 59.82  \\
                           & $\dcu$  & \textbf{43.06} & \textbf{29.64} & \textbf{39.39} & \textbf{60.15} & \textbf{58.35}  \\
\bottomrule
\end{tabular}
}
\end{center}
\end{table}
\setlength{\tabcolsep}{1.4pt}

\begin{figure}[tb]
\footnotesize
\centering
\setlength{\fboxrule}{0.1pt}%
\setlength{\fboxsep}{0pt}%
\begin{tabular}{@{}r@{\ }M{29mm}@{\ }M{29mm}@{\ \ }M{29mm}@{\ }M{29mm}@{\ }M{29mm}@{}}
   & \multicolumn{2}{c}{\footnotesize \textbf{RAFT} Perturbations} & \footnotesize Pert. Input $\img_t + \delta_t$ & \footnotesize Pert. Input $\img_{t+1}+\delta_{t+1}$ & \footnotesize Adversarial Flow $\fadv$ \\
$\dd$ & \includegraphics[width=29mm]{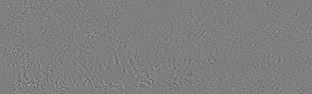} & \includegraphics[width=29mm]{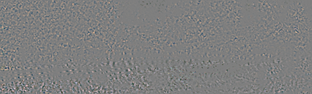} & \includegraphics[width=29mm]{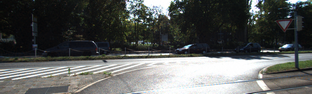} & \includegraphics[width=29mm]{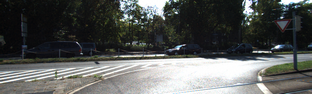} & \fcolorbox{gray}{white}{\includegraphics[width=29mm]{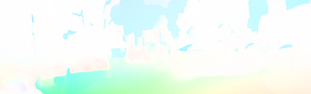}}\\
$\dc$ & \multicolumn{2}{M{60mm}}{\includegraphics[width=29mm]{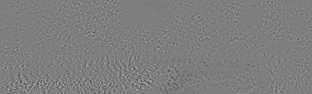}} & \includegraphics[width=29mm]{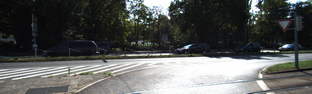} & \includegraphics[width=29mm]{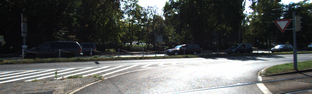} & \fcolorbox{gray}{white}{\includegraphics[width=29mm]{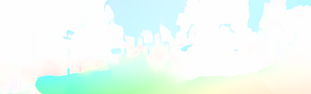}}\\
$\ddu$ & \includegraphics[width=29mm]{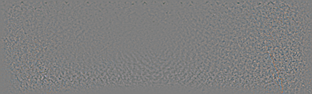} & \includegraphics[width=29mm]{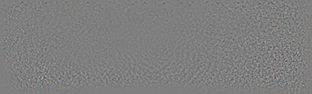} & \includegraphics[width=29mm]{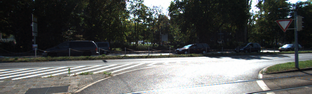} & \includegraphics[width=29mm]{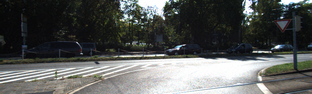} & \fcolorbox{gray}{white}{\includegraphics[width=29mm]{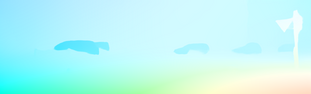}}\\
$\dcu$ & \multicolumn{2}{M{60mm}}{\includegraphics[width=29mm]{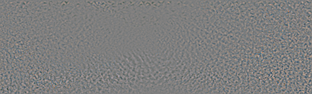}} & \includegraphics[width=29mm]{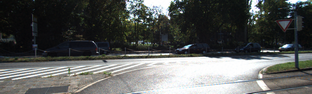} & \includegraphics[width=29mm]{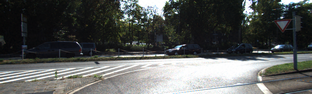} & \fcolorbox{gray}{white}{\includegraphics[width=29mm]{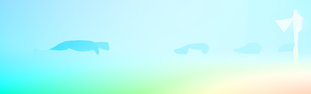}}\\[12pt]
   & \multicolumn{2}{c}{\footnotesize \textbf{SpyNet} Perturbations} & \footnotesize Pert. Input $\img_t + \delta_t$ & \footnotesize Pert. Input $\img_{t+1}+\delta_{t+1}$ & \footnotesize Adversarial Flow $\fadv$ \\
$\dd$ & \includegraphics[width=29mm]{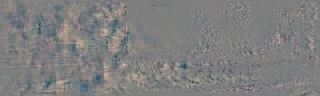} & \includegraphics[width=29mm]{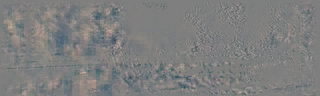} & \includegraphics[width=29mm]{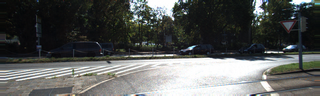} & \includegraphics[width=29mm]{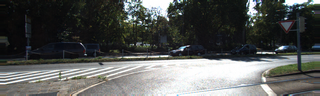} & \fcolorbox{gray}{white}{\includegraphics[width=29mm]{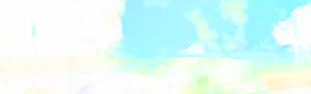}}\\
$\dc$ & \multicolumn{2}{M{60mm}}{\includegraphics[width=29mm]{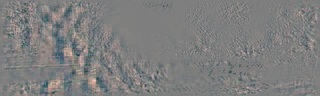}} & \includegraphics[width=29mm]{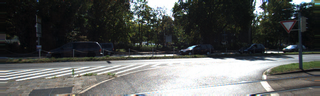} & \includegraphics[width=29mm]{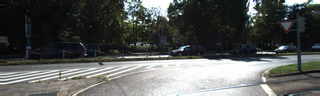} & \fcolorbox{gray}{white}{\includegraphics[width=29mm]{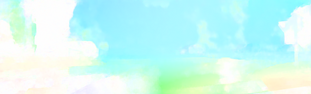}}\\
$\ddu$ & \includegraphics[width=29mm]{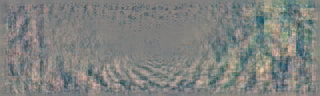} & \includegraphics[width=29mm]{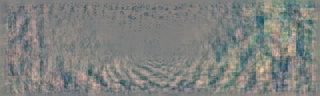} & \includegraphics[width=29mm]{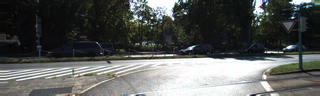} & \includegraphics[width=29mm]{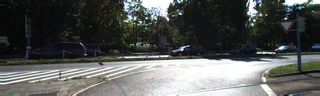} & \fcolorbox{gray}{white}{\includegraphics[width=29mm]{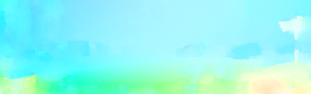}}\\
$\dcu$ & \multicolumn{2}{M{60mm}}{\includegraphics[width=29mm]{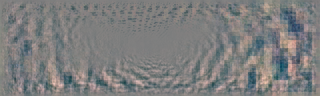}} & \includegraphics[width=29mm]{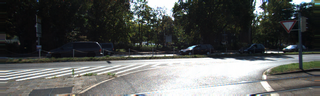} & \includegraphics[width=29mm]{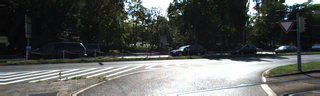} & \fcolorbox{gray}{white}{\includegraphics[width=29mm]{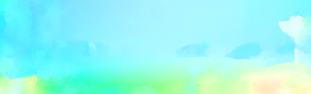}}\\
\end{tabular}
\caption{Different perturbations types (\emph{joint} and \emph{universal}) for \emph{RAFT} and \emph{SpyNet} on an exemplary KITTI frame pair, generated with PCFA ($\varepsilon_2 = 5\cdot 10^{-3}$, AEE loss, clipping box constraint).}
\label{fig:deltatypes_illustr}
\end{figure}

Fig.~\ref{fig:deltatypes_illustr} provides visualizations of the perturbations, perturbed inputs and adversarial flow for the zero-flow attack from Main Tab.~\ref{table:deltatype_kitti} on the networks RAFT and SpyNet.

\subsubsection{Transferability of Adversarial Perturbations}
\begin{figure}
\centering
\begin{tabular}{@{}M{10mm}@{\ \ \ }M{50mm}@{\ }M{50mm}@{}}
\multicolumn{1}{l}{\textbf{KITTI}} & \includegraphics[width=50mm]{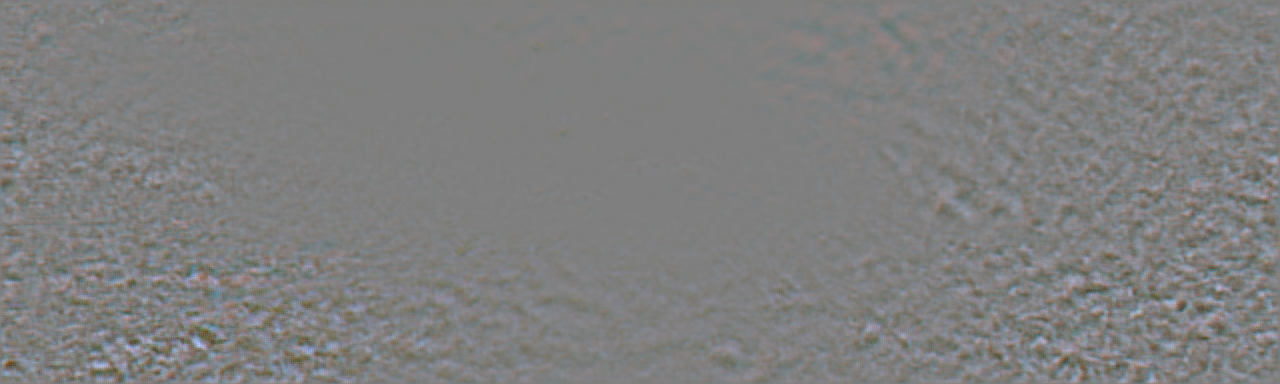} & \includegraphics[width=50mm]{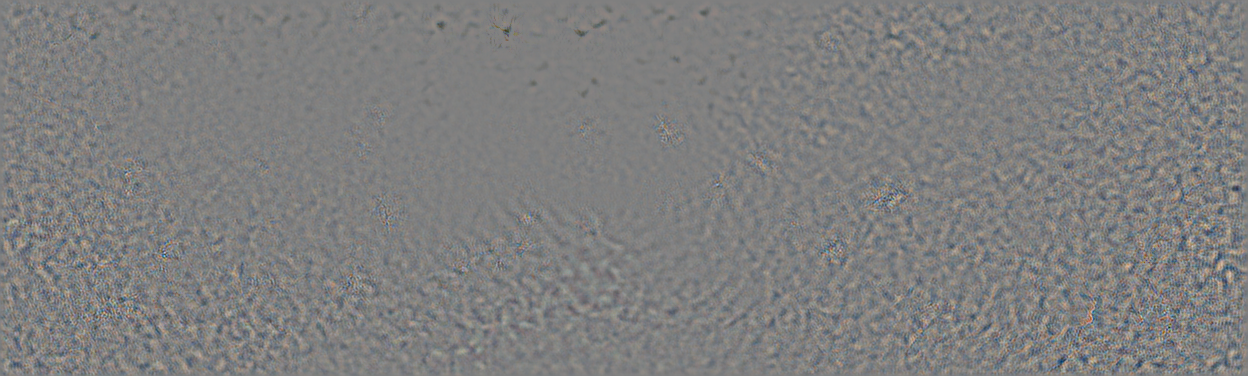}\\
\multicolumn{1}{l}{\textbf{Sintel}} & \includegraphics[width=50mm]{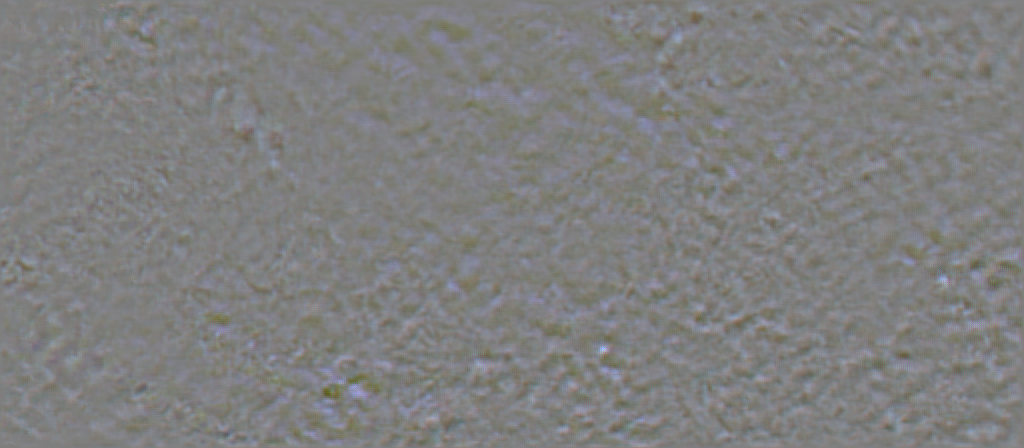} & \includegraphics[width=50mm]{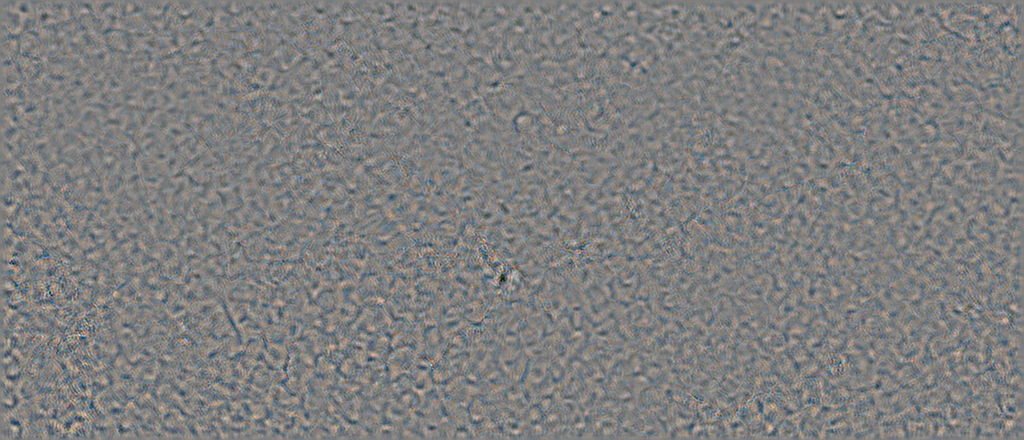}\\
& PWCNet~\cite{Sun2018PwcNetCnns} & GMA~\cite{Jiang2021LearningEstimateHidden}\\
\end{tabular}
\caption{Normalized \emph{universal perturbations} for different network architectures learned from the respective training datasets. Top row: KITTI. Bottom row: Sintel.}
\label{fig:universal_perturb_supp}
\end{figure}

Tab.~\ref{table:transfer_sintel} shows the transferability of samples for the Sintel final dataset, and complements Main Tab.~\ref{table:transfer_kitti}, which shows the same evaluation on KITTI.
Again, universal joint perturbations were trained on the training set for a specific network (top row), and then applied to the test set and tested on all available networks (first column).
On the Sintel dataset we observe clear trends among the networks in terms of robustness, where state-of-the-art networks like RAFT and GMA exhibit a consistent vulnerability to adversarial perturbations, while SpyNet's output is least distorted; irrespective of the network that was used to train the perturbation.
Fig.~\ref{fig:universal_perturb_supp} visualizes best universal perturbations for the Networks PWCNet and GMA, complementing the selection of networks from Main Fig~\ref{fig:universal_perturb}.
\setlength{\tabcolsep}{6pt}
\begin{table}
\begin{center}
\caption{Transferability of Sintel universal perturbations between \emph{training} and \emph{test} dataset and between different \emph{networks}, measured as adversarial robustness $\text{AEE}(\fadv, \flow)$. Large values denote a better transferability, smaller values indicate higher robustness.}
\label{table:transfer_sintel}
\begin{tabular}{l|@{\ \ \ }ccccc}
\toprule
\diagbox[width=7em, height=2em]{Eval.}{Train.}  & FlowNet2 & SpyNet & PWCNet & RAFT & GMA \\
\midrule
FlowNet2 &      \cellcolor{gray!25}3.11 &    2.54 &    2.47 &  1.35 & 1.28 \\
  SpyNet &      \underline{0.99} &    \cellcolor{gray!25}\underline{2.05} &    \underline{0.87} &  \underline{0.79} & \underline{0.72} \\
  PWCNet &      2.28 &    2.25 &    \cellcolor{gray!25}3.38 &  1.22 & 1.11 \\
    RAFT &      \textbf{7.78} &    \textbf{6.79} &    \textbf{6.86} &  \cellcolor{gray!25}\textbf{8.44} & \textbf{8.18} \\
     GMA &      6.77 &    5.72 &    5.79 &  7.28 & \cellcolor{gray!25}7.22 \\
\bottomrule
\end{tabular}
\end{center}
\end{table}
\setlength{\tabcolsep}{1.4pt}

\medskip
\noindent\textbf{Comparing the Patch Attack and PCFA.}
In Main Tab.~\ref{table:transfer_patchPCFA}, we compare the Patch Attack by Ranjan \etal~\cite{Ranjan2019AttackingOpticalFlow} and our PCFA in terms of distance between the original and perturbed prediction.
In both cases, the universal perturbations are trained on the Sintel final training set, and evaluated on test.
For the networks listed in the original publication, we use the patches from~\cite{Ranjan2019AttackingOpticalFlow}.
As RAFT and GMA are not included in the original publication, we use the official code with standard settings to generate them, i.e.\ a learning rate of $10^3$, $40$ epochs, $100$ images per epoch, two SGD steps per image.
In the following we discuss to what extend and under which assumptions the reported adversarial robustness numbers for PCFA and the Patch Attack~\cite{Ranjan2019AttackingOpticalFlow} are comparable.

\medskip
\noindent\textbf{Estimating the per-pixel \ltwo\ norm of the Patch Attack.}
To roughly estimate the per-pixel \ltwo\ norm for Patch Attack \cite{Ranjan2019AttackingOpticalFlow}, we use the following assumptions.
First, we assume the patch to introduce an additive distortion in the patch area, while adding zero in every location outside the patch $P$.
Further, we assume the patch is contained in the image area $P \subset I$.
And finally, we assume that the distortion adds a fixed value $\bar{b}$ to every location $p$ within the patch $P$ instead of individual values $b_p$.
Please note that this is a very conservative assumption, since among all additive distortions that have a mean absolute value $\bar{b}=\frac{1}{P}\sum_{p\in P} |b_p|$ within the patch $P$, the constant distortion $b_p=\bar{b}$ has the smallest \ltwo\ norm. Hence, in practice, the \ltwo\ norm of Patch Attack will be larger than our estimate.  The three aforementioned assumptions allow us to estimate the per-pixel \ltwo\ norm of the patch distortion as
\begin{align}
\varepsilon_2 &= \frac{\|\delta_P\|_2}{\sqrt{I}} 
    = \sqrt{\frac{\sum_{p\in I} b_p^2}{I}}
   =  \sqrt{\frac{\sum_{p\in P} b_p^2 + \sum_{p\notin P} 0^2}{I}}
   \;\;\;\stackrel{\mathclap{\substack{b_p = \bar{b}\\ p\in P}}}{=} 
   \;\;\sqrt{\frac{P \; \bar{b}^2}{I}}
   = \sqrt{\frac{P}{I}}|\bar{b}| \, . \label{equ:l2p_3}
\end{align}

While Eq.~\eqref{equ:l2p_3} refers to a patch in a single frame with a single channel, it generalizes in a straightforward manner to two patches in two frames with multiple channels per frame. To this end, one has to consider the joint norm $\|\delta_P^t, \delta_P^{t+1}\|_2$
and approximate it using the joint mean absolute value $|\bar{b}_{joint}|$ over all channels of both frames. Thus we obtain
\begin{align}
\varepsilon_2 &= \frac{\|\delta_t^P, \delta_{t+1}^P\|_2}{\sqrt{2IC}}
 = \sqrt{\frac{P}{I}}|\bar{b}_{joint}| \, . \label{equ:l2p_4}
\end{align}
Please note that computing the mean absolute value over all channels of both frames already performs a normalization by the number of frames and the number of channels. Hence, the factor does not change compared to Eq.~\eqref{equ:l2p_3}.

In our comparison to the patch-based method of Ranjan {\em et al.} \cite{Ranjan2019AttackingOpticalFlow} we consider a patch of \O $102$ pixels, which corresponds to approximately $8171$ pixels.
For a typical KITTI frame with a resolution of $I\!=\!1242$$\times$$375$, this results in a perturbation of about $1.75$\% of all pixels.
Given that the patches have colors which are rarely present in a typical KITTI frame, we conservatively estimate that the average additive perturbation $\bar{b}_{joint}$  per patch is about $3-30$\% of the valid color range.
With Eq.~\eqref{equ:l2p_4} this translates to an average color change over the whole frame of $0.40-3.97$\%.
We compare the Patch Attack to PCFA with an \ltwo\ bound of $\varepsilon_2 = 5\cdot 10^{-3}$, which translates to an average change of $0.50$\% of the color range per pixel.
This is at the lower end of our conservatively estimated range for the Patch Attack.
Provided that comparing two methods with different underlying concepts is difficult, we aimed for a comparison that is as fair as possible under the methodological constraints.

\putbib[bibliography]
\end{bibunit}

\end{document}

%% file: Images/EpsPlot_predadv-target_CWFlow_FGSM_flow_zero_clipping_Kitti15.pdf_tex
%% Creator: Inkscape 1.1.2 (1:1.1+202202050950+0a00cf5339), www.inkscape.org
%% Accompanies image file 'EpsPlot_predadv-target_CWFlow_FGSM_flow_zero_clipping_Kitti15.pdf' (pdf, eps, ps)
\begingroup%
  \makeatletter%
  \providecommand\color[2][]{%
    \errmessage{(Inkscape) Color is used for the text in Inkscape, but the package 'color.sty' is not loaded}%
    \renewcommand\color[2][]{}%
  }%
  \providecommand\transparent[1]{%
    \errmessage{(Inkscape) Transparency is used (non-zero) for the text in Inkscape, but the package 'transparent.sty' is not loaded}%
    \renewcommand\transparent[1]{}%
  }%
  \providecommand\rotatebox[2]{#2}%
  \newcommand*\fsize{\dimexpr\f@size pt\relax}%
  \newcommand*\lineheight[1]{\fontsize{\fsize}{#1\fsize}\selectfont}%
  \ifx\svgwidth\undefined%
    \setlength{\unitlength}{376.271181bp}%
    \ifx\svgscale\undefined%
      \relax%
    \else%
      \setlength{\unitlength}{\unitlength * \real{\svgscale}}%
    \fi%
  \else%
    \setlength{\unitlength}{\svgwidth}%
  \fi%
  \global\let\svgwidth\undefined%
  \global\let\svgscale\undefined%
  \makeatother%
  \begin{picture}(1,0.46556568)%
    \lineheight{1}%
    \setlength\tabcolsep{0pt}%
    \put(0,0){\includegraphics[width=\unitlength,page=1]{EpsPlot_predadv-target_CWFlow_FGSM_flow_zero_clipping_Kitti15.pdf}}%
  \end{picture}%
\endgroup%

%% file: Images/RobQual_pred-predadv_CWFlow_0.005_zero_change_of_variables_Kitti15_bm.pdf_tex
%% Creator: Inkscape 1.1.2 (1:1.1+202202050950+0a00cf5339), www.inkscape.org
%% Accompanies image file 'RobQual_pred-predadv_CWFlow_0.005_zero_change_of_variables_Kitti15_bm.pdf' (pdf, eps, ps)
\begingroup%
  \makeatletter%
  \providecommand\color[2][]{%
    \errmessage{(Inkscape) Color is used for the text in Inkscape, but the package 'color.sty' is not loaded}%
    \renewcommand\color[2][]{}%
  }%
  \providecommand\transparent[1]{%
    \errmessage{(Inkscape) Transparency is used (non-zero) for the text in Inkscape, but the package 'transparent.sty' is not loaded}%
    \renewcommand\transparent[1]{}%
  }%
  \providecommand\rotatebox[2]{#2}%
  \newcommand*\fsize{\dimexpr\f@size pt\relax}%
  \newcommand*\lineheight[1]{\fontsize{\fsize}{#1\fsize}\selectfont}%
  \ifx\svgwidth\undefined%
    \setlength{\unitlength}{197.997875bp}%
    \ifx\svgscale\undefined%
      \relax%
    \else%
      \setlength{\unitlength}{\unitlength * \real{\svgscale}}%
    \fi%
  \else%
    \setlength{\unitlength}{\svgwidth}%
  \fi%
  \global\let\svgwidth\undefined%
  \global\let\svgscale\undefined%
  \makeatother%
  \begin{picture}(1,0.67144011)%
    \lineheight{1}%
    \setlength\tabcolsep{0pt}%
    \put(0,0){\includegraphics[width=\unitlength,page=1]{RobQual_pred-predadv_CWFlow_0.005_zero_change_of_variables_Kitti15_bm.pdf}}%
  \end{picture}%
\endgroup%

%% file: Images/RobQual_pred-predadv_CWFlow_0.005_zero_change_of_variables_Sintel_bm.pdf_tex
%% Creator: Inkscape 1.1.2 (1:1.1+202202050950+0a00cf5339), www.inkscape.org
%% Accompanies image file 'RobQual_pred-predadv_CWFlow_0.005_zero_change_of_variables_Sintel_bm.pdf' (pdf, eps, ps)
\begingroup%
  \makeatletter%
  \providecommand\color[2][]{%
    \errmessage{(Inkscape) Color is used for the text in Inkscape, but the package 'color.sty' is not loaded}%
    \renewcommand\color[2][]{}%
  }%
  \providecommand\transparent[1]{%
    \errmessage{(Inkscape) Transparency is used (non-zero) for the text in Inkscape, but the package 'transparent.sty' is not loaded}%
    \renewcommand\transparent[1]{}%
  }%
  \providecommand\rotatebox[2]{#2}%
  \newcommand*\fsize{\dimexpr\f@size pt\relax}%
  \newcommand*\lineheight[1]{\fontsize{\fsize}{#1\fsize}\selectfont}%
  \ifx\svgwidth\undefined%
    \setlength{\unitlength}{194.123503bp}%
    \ifx\svgscale\undefined%
      \relax%
    \else%
      \setlength{\unitlength}{\unitlength * \real{\svgscale}}%
    \fi%
  \else%
    \setlength{\unitlength}{\svgwidth}%
  \fi%
  \global\let\svgwidth\undefined%
  \global\let\svgscale\undefined%
  \makeatother%
  \begin{picture}(1,0.6848409)%
    \lineheight{1}%
    \setlength\tabcolsep{0pt}%
    \put(0,0){\includegraphics[width=\unitlength,page=1]{RobQual_pred-predadv_CWFlow_0.005_zero_change_of_variables_Sintel_bm.pdf}}%
  \end{picture}%
\endgroup%

%% file: Images/EpsPlot_pred-predadv_CWFlow_FGSM_flow_zero_clipping_Kitti15.pdf_tex
%% Creator: Inkscape 1.1.2 (1:1.1+202202050950+0a00cf5339), www.inkscape.org
%% Accompanies image file 'EpsPlot_pred-predadv_CWFlow_FGSM_flow_zero_clipping_Kitti15.pdf' (pdf, eps, ps)
\begingroup%
  \makeatletter%
  \providecommand\color[2][]{%
    \errmessage{(Inkscape) Color is used for the text in Inkscape, but the package 'color.sty' is not loaded}%
    \renewcommand\color[2][]{}%
  }%
  \providecommand\transparent[1]{%
    \errmessage{(Inkscape) Transparency is used (non-zero) for the text in Inkscape, but the package 'transparent.sty' is not loaded}%
    \renewcommand\transparent[1]{}%
  }%
  \providecommand\rotatebox[2]{#2}%
  \newcommand*\fsize{\dimexpr\f@size pt\relax}%
  \newcommand*\lineheight[1]{\fontsize{\fsize}{#1\fsize}\selectfont}%
  \ifx\svgwidth\undefined%
    \setlength{\unitlength}{376.271181bp}%
    \ifx\svgscale\undefined%
      \relax%
    \else%
      \setlength{\unitlength}{\unitlength * \real{\svgscale}}%
    \fi%
  \else%
    \setlength{\unitlength}{\svgwidth}%
  \fi%
  \global\let\svgwidth\undefined%
  \global\let\svgscale\undefined%
  \makeatother%
  \begin{picture}(1,0.49140853)%
    \lineheight{1}%
    \setlength\tabcolsep{0pt}%
    \put(0,0){\includegraphics[width=\unitlength,page=1]{EpsPlot_pred-predadv_CWFlow_FGSM_flow_zero_clipping_Kitti15.pdf}}%
  \end{picture}%
\endgroup%

%% file: templateArxiv.bbl
\begin{thebibliography}{10}

\bibitem{Ilg2017Flownet2Evolution}
Eddy Ilg, Nikolaus Mayer, Tonmoy Saikia, Margret Keuper, Alexey Dosovitskiy,
  and Thomas Brox.
\newblock {FlowNet} 2.0: Evolution of optical flow estimation with deep
  networks.
\newblock In {\em Proc. IEEE/CVF Conference on Computer Vision and Pattern
  Recognition (CVPR)}, 2017.

\bibitem{Reda2017Flownet2PytorchPytorch}
Fitsum Reda, Robert Pottorff, Jon Barker, and Bryan Catanzaro.
\newblock flownet2-pytorch: Pytorch implementation of {FlowNet} 2.0: Evolution
  of optical flow estimation with deep networks, 2017.

\bibitem{Sun2018PwcNetCnns}
Deqing Sun, Xiaodong Yang, Ming-Yu Liu, and Jan Kautz.
\newblock {PWC-Net}: {CNNs} for optical flow using pyramid, warping, and cost
  volume.
\newblock In {\em Proc. IEEE/CVF Conference on Computer Vision and Pattern
  Recognition (CVPR)}, 2018.

\bibitem{Ranjan2017OpticalFlowEstimation}
Anurag Ranjan and Michael~J. Black.
\newblock Optical flow estimation using a spatial pyramid network.
\newblock In {\em Proc. IEEE/CVF Conference on Computer Vision and Pattern
  Recognition (CVPR)}, 2017.

\bibitem{Niklaus2018PytorchSpyNet}
Simon Niklaus.
\newblock A reimplementation of {SPyNet} using {PyTorch}, 2018.

\bibitem{teed2020raft}
Zachary Teed and Jia Deng.
\newblock {RAFT}: Recurrent all-pairs field transforms for optical flow.
\newblock In {\em Proc. European Conference on Computer Vision (ECCV)}, pages
  402--419, 2020.

\bibitem{Jiang2021LearningEstimateHidden}
Shihao Jiang, Dylan Campbell, Yao Lu, Hongdong Li, and Richard Hartley.
\newblock Learning to estimate hidden motions with global motion aggregation.
\newblock In {\em Proc. IEEE/CVF International Conference on Computer Vision
  (ICCV)}, pages 9772--9781, 2021.

\bibitem{Menze2015Joint3dEstimation}
Moritz Menze, Christian Heipke, and Andreas Geiger.
\newblock Joint {3D} estimation of vehicles and scene flow.
\newblock In {\em Proc. ISPRS Workshop on Image Sequence Analysis (ISA)}, 2015.

\bibitem{Butler2012NaturalisticOpenSource}
D.~J. Butler, J.~Wulff, G.~B. Stanley, and M.~J. Black.
\newblock A naturalistic open source movie for optical flow evaluation.
\newblock In {\em Proc. European Conference on Computer Vision (ECCV)}, pages
  611--625, 2012.

\bibitem{Schrodi2022TowardsUnderstandingAdversarial}
Simon Schrodi, Tonmoy Saikia, and Thomas Brox.
\newblock Towards understanding adversarial robustness of optical flow
  networks.
\newblock In {\em Proc. IEEE/CVF Conference on Computer Vision and Pattern
  Recognition (CVPR)}, pages 8916--8924, 2022.

\bibitem{Kurakin2017AdversarialMachineLearning}
Alexey Kurakin, Ian Goodfellow, and Samy Bengio.
\newblock Adversarial machine learning at scale.
\newblock In {\em arXiv preprint 1611.01236}, 2017.

\bibitem{Ranjan2019AttackingOpticalFlow}
Anurag Ranjan, Joel Janai, Andreas Geiger, and Michael~J. Black.
\newblock Attacking optical flow.
\newblock In {\em Proc. IEEE/CVF International Conference on Computer Vision
  (ICCV)}, 2019.

\end{thebibliography}


\begin{thebibliography}{10}

\bibitem{teed2020raft}
Zachary Teed and Jia Deng.
\newblock {RAFT}: Recurrent all-pairs field transforms for optical flow.
\newblock In {\em Proc. European Conference on Computer Vision (ECCV)}, pages
  402--419, 2020.

\bibitem{Ullah2019ActionReconition}
Amin Ullah, Khan Muhammad, Javier Del~Ser, Sung~Wook Baik, and Victor Hugo~C.
  de~Albuquerque.
\newblock Activity recognition using temporal optical flow convolutional
  features and multilayer lstm.
\newblock {\em IEEE Transactions on Industrial Electronics (TIE)},
  66(12):9692--9702, 2019.

\bibitem{Wang2020Superresolution}
L.~Wang, Y.~Guo, L.~Liu, Z.~Lin, X.~Deng, and W.~An.
\newblock Deep video super-resolution using {HR} optical flow estimation.
\newblock {\em IEEE Transactions on Image Processing (TIP)}, 29:4323--4336,
  2020.

\bibitem{Zhang2020Robotics}
T.~Zhang, H.~Zhang, Y.~Li, Y.~Nakamura, and L.~Zhang.
\newblock Flowfusion: Dynamic dense {RGB-D} {SLAM} based on optical flow.
\newblock In {\em Proc, IEEE International Conference on Robotics and
  Automation (ICRA)}, pages 7322--7328, 2020.

\bibitem{Stegmaier2020DifferencesEpidemicSpread}
Tabea Stegmaier, Eva Oellingrath, Mirko Himmel, and Simon Fraas.
\newblock Differences in epidemic spread patterns of norovirus and influenza
  seasons of {Germany}: an application of optical flow analysis in
  epidemiology.
\newblock {\em NatureResearch Scientific Reports}, 10(1):1--14, 2020.

\bibitem{Horn1981HornSchunck}
Berthold K.~P. Horn and Brian~G. Schunck.
\newblock Determining optical flow.
\newblock {\em Artificial Intelligence (AI)}, 17(1-3):185--203, 1981.

\bibitem{Black1993FrameworkRobustEstimation}
M.~J. Black and P.~Anandan.
\newblock A framework for the robust estimation of optical flow.
\newblock In {\em Proc. IEEE International Conference on Computer Vision
  (ICCV)}, pages 231--236, 1993.

\bibitem{Brox2004Warping}
T.~Brox, A.~Bruhn, N.~Papenberg, and J.~Weickert.
\newblock High accuracy optical flow estimation based on a theory for warping.
\newblock In {\em Proc. European Conference on Computer Vision (ECCV)}, pages
  25--36, 2004.

\bibitem{Brox2010LDOF}
T.~Brox and J.~Malik.
\newblock Large displacement optical flow: Descriptor matching in variational
  motion estimation.
\newblock {\em IEEE Transactions on Pattern Analysis and Machine Intelligence
  (PAMI)}, 33(3):500--513, 2011.

\bibitem{Sun2010Secrets}
Deqing Sun, Stefan Roth, and Michael Black.
\newblock Secrets of optical flow estimation and their principles.
\newblock In {\em Proc. IEEE Conference on Computer Vision and Pattern
  Recognition (CVPR)}, pages 2432--2499, 2010.

\bibitem{Ilg2017Flownet2Evolution}
Eddy Ilg, Nikolaus Mayer, Tonmoy Saikia, Margret Keuper, Alexey Dosovitskiy,
  and Thomas Brox.
\newblock {FlowNet} 2.0: Evolution of optical flow estimation with deep
  networks.
\newblock In {\em Proc. IEEE/CVF Conference on Computer Vision and Pattern
  Recognition (CVPR)}, 2017.

\bibitem{Ranjan2017OpticalFlowEstimation}
Anurag Ranjan and Michael~J. Black.
\newblock Optical flow estimation using a spatial pyramid network.
\newblock In {\em Proc. IEEE/CVF Conference on Computer Vision and Pattern
  Recognition (CVPR)}, 2017.

\bibitem{Sun2018PwcNetCnns}
Deqing Sun, Xiaodong Yang, Ming-Yu Liu, and Jan Kautz.
\newblock {PWC-Net}: {CNNs} for optical flow using pyramid, warping, and cost
  volume.
\newblock In {\em Proc. IEEE/CVF Conference on Computer Vision and Pattern
  Recognition (CVPR)}, 2018.

\bibitem{Yang2019_VCN}
G.~Yang and D.~Ramanan.
\newblock Volumetric correspondence networks for optical flow.
\newblock In {\em Proc. Conference on Neural Information Processing Systems
  (NeurIPS)}, pages 794--805, 2019.

\bibitem{Yin2019_HD3}
Z.~Yin, T.~Darrell, and F.~Yu.
\newblock Hierarchical discrete distribution decomposition for match density
  estimation.
\newblock In {\em Proc. IEEE/CVF Conference on Computer Vision and Pattern
  Recognition (CVPR)}, pages 6044--6053, 2019.

\bibitem{Jiang2021LearningEstimateHidden}
Shihao Jiang, Dylan Campbell, Yao Lu, Hongdong Li, and Richard Hartley.
\newblock Learning to estimate hidden motions with global motion aggregation.
\newblock In {\em Proc. IEEE/CVF International Conference on Computer Vision
  (ICCV)}, pages 9772--9781, 2021.

\bibitem{Zhang2021_Separable}
F.~Zhang, O.~Woodford, V.~Prisacariu, and P.~Torr.
\newblock Separable flow: Learning motion cost volumes for optical flow
  estimation.
\newblock In {\em Proc. IEEE/CVF International Conference on Computer Vision
  (ICCV)}, pages 10807--10817, 2021.

\bibitem{Barron1994Performance}
J.~L. Barron, D.~J. Fleet, and S.~S. Beauchemin.
\newblock Performance of optical flow techniques.
\newblock {\em International Journal of Computer Vision (IJCV)}, 12:43--77,
  1994.

\bibitem{Baker2011Middlebury}
S.~Baker, D.~Scharstein, J.~P. Lewis, S.~Roth, M.~J. Black, and R.~Szeliski.
\newblock A database and evaluation methodology for optical flow.
\newblock {\em International Journal of Computer Vision (IJCV)}, 92(1):1--31,
  2011.

\bibitem{Geiger2012AreWeReady}
Andreas Geiger, Philip Lenz, and Raquel Urtasun.
\newblock Are we ready for autonomous driving? {T}he {KITTI} vision benchmark
  suite.
\newblock In {\em Proc. IEEE Conference on Computer Vision and Pattern
  Recognition (CVPR)}, 2012.

\bibitem{Butler2012NaturalisticOpenSource}
D.~J. Butler, J.~Wulff, G.~B. Stanley, and M.~J. Black.
\newblock A naturalistic open source movie for optical flow evaluation.
\newblock In {\em Proc. European Conference on Computer Vision (ECCV)}, pages
  611--625, 2012.

\bibitem{Menze2015Joint3dEstimation}
Moritz Menze, Christian Heipke, and Andreas Geiger.
\newblock Joint {3D} estimation of vehicles and scene flow.
\newblock In {\em Proc. ISPRS Workshop on Image Sequence Analysis (ISA)}, 2015.

\bibitem{Yu2020CardiacMotion}
H.~Yu, X.~Chen, H.~Shi, T.~Chen, T.~S. Huang, and S.~Sun.
\newblock Motion pyramid networks for accurate and efficient cardiac motion
  estimation.
\newblock In {\em Proc. International Conference on Medical Image Computing and
  Computer-Assisted Intervention (MICCAI)}, pages 436--446, 2020.

\bibitem{Tehrani2020Ultrasound}
Ali Tehrani, Morteza Mirzae, and Hassan Rivaz.
\newblock Semi-supervised training of optical flow convolutional neural
  networks in ultrasound elastography.
\newblock In {\em Proc. International Conference on Medical Image Computing and
  Computer-Assisted Intervention (MICCAI)}, pages 504--513, 2020.

\bibitem{Capito2020OpticalFlowBased}
Linda Capito, Umit Ozguner, and Keith Redmill.
\newblock Optical flow based visual potential field for autonomous driving.
\newblock In {\em IEEE Intelligent Vehicles Symposium (IV)}, pages 885--891,
  2020.

\bibitem{Wang2021EndEndInteractive}
Hengli Wang, Peide Cai, Rui Fan, Yuxiang Sun, and Ming Liu.
\newblock End-to-end interactive prediction and planning with optical flow
  distillation for autonomous driving.
\newblock In {\em Proc. IEEE/CVF Conference on Computer Vision and Pattern
  Recognition Workshops (CVPR-W)}, pages 2229--2238, 2021.

\bibitem{Weijer2004RobustOpticalFlow}
J.~{van de Weijer} and Th. Gevers.
\newblock Robust optical flow from photometric invariants.
\newblock In {\em Proc. IEEE International Conference on Image Processing
  (ICIP)}, volume~3, pages 1835--1838, 2004.

\bibitem{Stein2004Census}
F.~Stein.
\newblock Efficient computation of optical flow using the census transform.
\newblock In {\em Proc. German Conference on Pattern Recognition (DAGM)}, pages
  79--86, 2004.

\bibitem{Liu2010SIFTFlow}
Ce~Liu, Jenny Yuen, and Antonio Torralba.
\newblock {SIFT} flow: Dense correspondence across scenes and its applications.
\newblock {\em IEEE Transactions on Pattern Analysis and Machine Intelligence
  (PAMI)}, 33(5):978--994, 2010.

\bibitem{Li2018RobustOpticalFlow}
Ruoteng Li, Robby~T. Tan, and Loong-Fah Cheong.
\newblock Robust optical flow in rainy scenes.
\newblock In {\em Proc. European Conference on Computer Vision (ECCV)}, 2018.

\bibitem{Ranjan2019AttackingOpticalFlow}
Anurag Ranjan, Joel Janai, Andreas Geiger, and Michael~J. Black.
\newblock Attacking optical flow.
\newblock In {\em Proc. IEEE/CVF International Conference on Computer Vision
  (ICCV)}, 2019.

\bibitem{Schrodi2022TowardsUnderstandingAdversarial}
Simon Schrodi, Tonmoy Saikia, and Thomas Brox.
\newblock Towards understanding adversarial robustness of optical flow
  networks.
\newblock In {\em Proc. IEEE/CVF Conference on Computer Vision and Pattern
  Recognition (CVPR)}, pages 8916--8924, 2022.

\bibitem{Goodfellow2014ExplainingHarnessingAdversarial}
Ian~J. Goodfellow, Jonathon Shlens, and Christian Szegedy.
\newblock Explaining and harnessing adversarial examples.
\newblock In {\em arXiv preprint 1412.6572}, 2014.

\bibitem{Bruhn2005CLG}
A.~Bruhn, J.~Weickert, and C.~Schn\"orr.
\newblock Lucas/{K}anade meets {H}orn/{S}chunck: Combining local and global
  optic flow methods.
\newblock {\em International Journal of Computer Vision (IJCV)},
  61(3):211--231, 2005.

\bibitem{Szegedy2014IntriguingPropertiesNeural}
Christian Szegedy, Wojciech Zaremba, Ilya Sutskever, Joan Bruna, Dumitru Erhan,
  Ian Goodfellow, and Rob Fergus.
\newblock Intriguing properties of neural networks.
\newblock In {\em Proc. International Conference on Learning Representations
  (ICLR)}, 2014.

\bibitem{Virmaux2018LipschitzRegularityDeep}
Aladin Virmaux and Kevin Scaman.
\newblock Lipschitz regularity of deep neural networks: analysis and efficient
  estimation.
\newblock In {\em Proc. Conference on Neural Information Processing Systems
  (NeurIPS)}, 2018.

\bibitem{Carlini2017TowardsEvaluatingRobustness}
Nicholas Carlini and David Wagner.
\newblock Towards evaluating the robustness of neural networks.
\newblock In {\em Proc. IEEE Symposium on Security and Privacy (SP)}, pages
  39--57, 2017.

\bibitem{Kurakin2017AdversarialMachineLearning}
Alexey Kurakin, Ian Goodfellow, and Samy Bengio.
\newblock Adversarial machine learning at scale.
\newblock In {\em arXiv preprint 1611.01236}, 2017.

\bibitem{Wong2021StereopagnosiaFoolingStereo}
Alex Wong, Mukund Mundhra, and Stefano Soatto.
\newblock Stereopagnosia: Fooling stereo networks with adversarial
  perturbations.
\newblock {\em Proc. AAAI Conference on Artificial Intelligence (AAAI)},
  35(4):2879--2888, 2021.

\bibitem{Dong2018BoostingAdversarialAttacks}
Yinpeng Dong, Fangzhou Liao, Tianyu Pang, Hang Su, Jun Zhu, Xiaolin Hu, and
  Jianguo Li.
\newblock Boosting adversarial attacks with momentum.
\newblock In {\em Proc. IEEE/CVF Conference on Computer Vision and Pattern
  Recognition (CVPR)}, 2018.

\bibitem{Anand2020AdversarialPatchDefense}
Adithya~Prem Anand, H.~Gokul, Harish Srinivasan, Pranav Vijay, and Vineeth
  Vijayaraghavan.
\newblock Adversarial patch defense for optical flow networks in video action
  recognition.
\newblock In {\em Proc. IEEE International Conference on Machine Learning and
  Applications (ICMLA)}, pages 1289--1296, 2020.

\bibitem{Brown2018AdversarialPatch}
Tom~B. Brown, Dandelion Man\'{e}, Aurko Roy, Mart\'{i}n Abadi, and Justin
  Gilmer.
\newblock Adversarial patch.
\newblock In {\em arXiv preprint 1712.09665}, 2018.

\bibitem{MoosaviDezfooli2017UniversalAdversarialPerturbations}
Seyed-Mohsen Moosavi-Dezfooli, Alhussein Fawzi, Omar Fawzi, and Pascal
  Frossard.
\newblock Universal adversarial perturbations.
\newblock In {\em Proc. IEEE/CVF Conference on Computer Vision and Pattern
  Recognition (CVPR)}, 2017.

\bibitem{Shafahi2020UniversalAdversarialTraining}
Ali Shafahi, Mahyar Najibi, Zheng Xu, John Dickerson, Larry~S. Davis, and Tom
  Goldstein.
\newblock Universal adversarial training.
\newblock {\em Proc. AAAI Conference on Artificial Intelligence (AAAI)},
  34(04):5636--5643, 04 2020.

\bibitem{Deng2020UniversalAdversarialAttack}
Yingpeng Deng and Lina~J. Karam.
\newblock Universal adversarial attack via enhanced projected gradient descent.
\newblock In {\em Proc. IEEE International Conference on Image Processing
  (ICIP)}, pages 1241--1245, 2020.

\bibitem{Xu2020AdversarialAttacksDefenses}
Han Xu, Yao Ma, Hao-Chen Liu, Debayan Deb, Hui Liu, Ji-Liang Tang, and Anil~K.
  Jain.
\newblock Adversarial attacks and defenses in images, graphs and text: A
  review.
\newblock {\em International Journal of Automation and Computing (IJAC)},
  17(2):151--178, 2020.

\bibitem{Tsipras2019RobustnessMayBe}
Dimitris Tsipras, Shibani Santurkar, Logan Engstrom, Alexander Turner, and
  Aleksander Madry.
\newblock Robustness may be at odds with accuracy.
\newblock In {\em Proc. International Conference on Learning Representations
  (ICLR)}, 2019.

\bibitem{Nocedal2006NumericalOptimization}
Jorge Nocedal and Stephen~J. Wright.
\newblock {\em Numerical optimization}.
\newblock Springer, 2nd edition, 2006.

\bibitem{Nocedal1980UpdatingQuasiNewton}
Jorge Nocedal.
\newblock Updating quasi-{N}ewton matrices with limited storage.
\newblock {\em Mathematics of Computation}, 35(151):773--782, 1980.

\bibitem{Paszke2019Pytorch}
Adam Paszke, Sam Gross, Francisco Massa, Adam Lerer, James Bradbury, Gregory
  Chanan, Trevor Killeen, Zeming Lin, Natalia Gimelshein, Luca Antiga, Alban
  Desmaison, Andreas Kopf, Edward Yang, Zachary DeVito, Martin Raison, Alykhan
  Tejani, Sasank Chilamkurthy, Benoit Steiner, Lu~Fang, Junjie Bai, and Soumith
  Chintala.
\newblock {PyTorch}: An imperative style, high-performance deep learning
  library.
\newblock In {\em Proc. Conference on Neural Information Processing Systems
  (NeurIPS)}, pages 8024--8035, 2019.

\bibitem{Reda2017Flownet2PytorchPytorch}
Fitsum Reda, Robert Pottorff, Jon Barker, and Bryan Catanzaro.
\newblock flownet2-pytorch: Pytorch implementation of {FlowNet} 2.0: Evolution
  of optical flow estimation with deep networks, 2017.

\bibitem{Niklaus2018PytorchSpyNet}
Simon Niklaus.
\newblock A reimplementation of {SPyNet} using {PyTorch}, 2018.

\end{thebibliography}
